\documentclass[twoside,11pt,preprint]{article}

\usepackage{jmlr2e}

\usepackage{mathtools}
\usepackage{comment}
\usepackage{amsmath}
\usepackage{hyperref}
\usepackage{multicol}
\usepackage{multirow}
\usepackage{xcolor}
\usepackage{array}
\usepackage{verbatim}
\usepackage{array}
\usepackage{float}
\usepackage{enumitem}
\usepackage{siunitx}
\usepackage{calc}
\usepackage{tensor}
\usepackage{bm}
\usepackage[bb=boondox]{mathalfa}

\usepackage{comment}
\allowdisplaybreaks[1]

\newtheorem{claim}{Claim}[section]

\newcommand{\ba}{\bm{a}}
\newcommand{\bb}{\bm{b}}
\newcommand{\bc}{\bm{c}}
\newcommand{\be}{\bm{e}}
\newcommand{\bx}{\bm{x}}
\newcommand{\by}{\bm{y}}
\newcommand{\bz}{\bm{z}}

\newcommand{\bs}{\bm{s}}

\newcommand{\N}{\mathbb{N}}

\newcommand{\R}{\mathbb{R}}
\newcommand{\C}{\mathbb{C}}

\newcommand{\ZZ}{\mathbb{Z}_{\geq 0}}

\newcommand{\sgn}{\text{sgn}}
\newcommand{\dimension}{\text{dim}}
\newcommand{\tr}{\text{tr}}
\newcommand{\alt}{\text{Alt}}

\newcommand{\calA}{\mathcal{A}}
\newcommand{\calC}{\mathcal{C}}
\newcommand{\calS}{\mathcal{S}}
\newcommand{\calO}{\mathcal{O}}
\newcommand{\calI}{\mathcal{I}}

\ShortHeadings{Uniform $C^k$-Approximations for Symmetric \& Antisymmetric Functions}{Ganguly, Sarkar and Tran}
\firstpageno{1}

\begin{document}

\title{\texorpdfstring{Uniform $\calC^k$}{Ck} Approximation of $G$-Invariant and Antisymmetric Functions, Embedding Dimensions, and Polynomial Representations}

\author{\name Soumya Ganguly             \email s1gangul@ucsd.edu \\
        \name Khoa Tran \email k2tran@ucsd.edu \\
       \addr Department of Mathematics\\
       University of California San Diego, \\
       San Diego, CA 92093, USA
       \AND
       \name Rahul Sarkar \email rsarkar@stanford.edu \\
       \addr Institute for Computational \& Mathematical Engineering\\
       Stanford University,\\
       Stanford, CA 94306, USA
}


\maketitle

\begin{abstract}
For any subgroup $G$ of the symmetric group $\mathcal{S}_n$ on $n$ symbols, we present results for the uniform $\calC^k$ approximation of $G$-invariant functions by $G$-invariant polynomials. For the case of totally symmetric functions ($G = \mathcal{S}_n$), we show that this gives rise to the \textit{sum-decomposition} Deep Sets ansatz of \citet{zaheer2018deepsets}, where both the inner and outer functions can be chosen to be smooth, and moreover, the inner function can be chosen to be independent of the target function being approximated. In particular, we show that the \textit{embedding dimension} required is independent of the regularity of the target function, the accuracy of the desired approximation, as well as $k$. Next, we show that a similar procedure allows us to obtain a uniform $\calC^{k}$ approximation of antisymmetric functions as a sum of $K$ terms, where each term is a product of a smooth totally symmetric function and a smooth antisymmetric homogeneous polynomial of degree at most $\binom{n}{2}$. We also provide upper and lower bounds on $K$ and show that $K$ is independent of the regularity of the target function, the desired approximation accuracy, and $k$.
\end{abstract}



\section{Introduction}\label{intro}

Currently, deep learning is widely used with great success in many applications, which have been organized into five broad categories: classification, localization, detection, segmentation, and registration \citep{Alzubaidi2021review}. Subsets of the applications are used for image processing \citep{krizhevsky2012imagenet, tian2020evolutionary}, audio and speech recognition \citep{Deng2013speech&signal, tjandra2017speech, Khurana2021speech}, natural language processing \citep{Deng2018book, otter2021languageSurvey, lauriola2022language}, autonomous vehicles \citep{Wan2021AV, Zhu2022AV}, and many others \citep{goodfellow2014numberRecognition, Silver2017gameOfGo, soffer2019radiology, Adeel2020AVspeech, Muhammad2021brainTumor, Zeleznik2021cardiovascular}. The common approach to most of these designs is building large neural networks (NNs) with deep layers; unfortunately, this often leads to one of the biggest challenges in deep learning in terms of computations, known as the curse of dimensionality. With large designs, the dimension of the data parameters increases; this causes an exponential increase in the number of necessary data samples that the model needs to properly learn the dataset. As a consequence, computational complexities also increase exponentially. 

Then, how does deep learning tackle the curse of dimensionality? There are many developing theories and methods that mitigate the problems of the curse, but the means to overcome the curse of dimensionality remains an open problem; there is also an assumption that the problem cannot be eliminated entirely due to the nature of neural networks, which is ubiquitous in deep learning. The current theories that suggest a large moderation of the curse include automated feature extractions \citep{laird1994automated} and the manifold hypothesis \citep{cayton2005algorithms}. Regularization methodologies such as dropout, batch normalization, and weight decay \citep{Garbin2020dropoutBatchN, loshchilov2018decoupled, xie2022understanding}, which serve to avert overfitting, can indirectly alleviate the curse by allowing the model to prevent learning noises in data. Another approach is exploiting the structure of the model function that can exhibit locality \citep{espi2015locality, zhang2021locality} and/or symmetries \citep{zaheer2018deepsets, pointnetqi2017, qi2017pointnetplpl}. The latter will be the main focus of our work; we propose theories that are relevant and favorable to deep learning problems with symmetries, specifically those involving invariance and antisymmetry. In fact, there has been a recent conjecture that under certain circumstances, permutation invariant NNs will have no barrier in linear interpolation of stochastic gradient descent solutions \citep{entezari2022role}.

In this paper, we study $\calC^k$ approximations and polynomial representations of $G$-invariant functions (for some [Lie] group $G$) and antisymmetric functions; note that $G$-invariant functions are symmetric (or permutation invariant) when $G$ is the symmetric group on the domain of the function. Intuitively, these symmetries allow the neural networks to learn from the inputs regardless of transformations by the action of $G$ for $G$-invariance or of transformations up to a signature for antisymmetry. There are two main advantages that come from our theories: First, the $\calC^k$ approximations give high accuracy, and they are useful for applications that require higher-order derivatives, e.g. solving many-electron Schrödinger equations \citep{Han_2019, Pfau_2020, Choo_2020}. Second, the polynomial representations are prevalent in STEM and bypass additional costs to have polynomial representations when the learning model only trains for some (continuous) functions. In fact, our work is a specific contribution to the long standing result called Universal Approximation Theorem, which informally means that any function can be approximated using neural networks. The result was originally proven by \citet{Cybenko1989ApproximationBS} for sigmoidal activation functions; \citet{hornik1991approximation} later proved this result for any-nonlinear activation functions, which has a similar proof to \citeauthor{Barron1993}'s Theorem on approximations of nonlinear, continuous functions. Due to this theorem, it is always sufficient to approximately represent $G$-invariant and antisymmetric functions instead of using exact representations. 

Furthermore, we specifically study approximations of $G$-invariant, symmetric, and antisymmetric functions because of their ubiquity in science and technology: Neural networks that are inherently invariant under an action $G$ include $n$-dimensional Convolutional Neural Networks (CNN) under translation, Spherical CNN under rotation, intrinsic CNN under the isometric group on the domains, and so on \citep{bronstein2021geometric}. These architectures are currently used in signal identification, object detection, image classification and segmentation, face recognition, etc \citep{Li2022CNNsurvey}. Similarly, some architectures with permutation invariance are Graph Neural Networks (GNN), Deep Sets \citep{zaheer2018deepsets}, and Transformer: Some applications of GNN are traffic forecasting, molecular optimization, rumor detection, node and graph classification, and many others \citep{Asif2021GNNsurvey}; of Deep Sets include point clouds prediction and bounding boxes \citep{soelch2019DeepSetLearning}; of transformer comprise of answer selection \citep{shao2019transformer} and stock volatility \citep{ramosperez2021multitransformer}. Lastly, neural networks can also learn solutions of systems that exhibit antisymmetry; in the physical sciences, there are many interests to find approximations of solutions to systems of many-fermion or many-Boson \citep{Choo_2020, Han_2019, Hermann_2020, Klus_2021, Luo_2019, Pfau_2020, Stokes_2020}. In particular, \citet{Pfau_2020} developed FermiNet as an ansatz on top of the variational Monte Carlo (VMC) model to approximate the solutions for many-electron systems; the ansatz is based on the notion of generalized Slater determinants \citep{hutter2020antisym}. The method gave large improvements on the VMC model for many atoms and small molecules; this essentially opens the door to solve previously intractable many-electron systems.

Finally, this study is made on the foundational work of \citet{zaheer2018deepsets}, who designed models with machine learning tasks defined on sets; the work was done on permutation invariant sets and equivariant tasks. Their work demonstrates great qualitative and quantitative results from experiments on statistic estimation, point cloud classification, set expansion, and outlier detection. Afterwards, many studies follow their work including PointNet \citep{pointnetqi2017, qi2017pointnetplpl}, Deep Potential \citep{zhang2018deepPot, zhang2018endtoend}, Set Aggregation Networks \citep{maziarka2019SAN}, and so on. To build the context of our work in relation to \textit{Deep Sets}, the technical and historical aspects are reserved in Section \ref{ssec:background}.

\subsection{Background} 
\label{ssec:background}

The main theorems of this paper concern the $\calC^k$ approximations and representations of $G$-invariant functions, totally symmetric functions, and $n$-antisymmetric functions. Let us define them here. First for convenience, call $d$ the \textit{space dimension} and $n$ the \textit{multivariate dimension}, and let $[n] = \{1, 2, \ldots, n\}$. By convention, $\N = \{ 1, 2, \ldots \}$, so we write non-negative numbers as $\ZZ = \N \cup \{0\}$. Throughout this study, we assume $n \geq 2$ and $d \geq 1$ unless stated otherwise. We also denote points in $\R^d$ using boldface letters such as $\bx_i$, and its components as $(x_{i1}, x_{i2}, \ldots, x_{id})$.

\begin{definition}[$G$-invariant function]
\label{def:G-invariantfn}
     Let $\Omega \subset \R^d$ and $G$ be a group which acts on $[n]$. A function  $f : \Omega^n \to \R$ is \textit{$G$-invariant} if 
    \begin{equation}
    \label{eq:totally_symmetric}
        f(\sigma . \bx) 
        \coloneqq f(\bx_{\sigma(1)}, \bx_{\sigma(2)}, \ldots, \bx_{\sigma(n)}) 
        = f(\bx_1, \bx_2, \ldots, \bx_n) 
        \coloneqq f(\bx),
    \end{equation}
    for all $\sigma \in G$ and $\{ \bx_i \}_{i = 1}^n \subset \Omega$.
\end{definition}

\begin{definition}[Totally symmetric function]
\label{def:totally_symmetric}
     A $G$-invariant function $f$ is called totally symmetric when $G = \calS_n$, the symmetric group.
\end{definition}

\begin{definition}[$n$-antisymmetric function]
\label{def:antisymmetric}
     Let $\Omega \subset \R^d$. A function $f : \Omega^n \to \R$ is called $n$-antisymmetric if 
    \begin{equation}
    \label{eq:antisymmetric}
        f(\sigma . \bx) 
        = \sgn(\sigma) f(\bx),
    \end{equation}
    for all $\sigma \in \calS_n$ and $\{ \bx_i \}_{i = 1}^n \subset \Omega$, where $\sgn(\sigma)$ is the sign of the permutation $\sigma$.
\end{definition}

As mentioned in the last section, representation of totally symmetric, $G$-invariant or $n$-antisymmetric functions using simpler such functions can significantly reduce the curse of dimensionality. There are results which represent these functions exactly and there are also many approximate representation theorems (some mentioned below). From a mathematical viewpoint, finding an exact representation is solicited but because of the intrinsic approximating nature of Neural networks (Universal approximation theorem), an approximate model works just as well for all practical purposes. In fact many times, keeping room for approximation allows us to get a much simpler representation- an example of which can be found in this current work. As mentioned before, this line of research boomed due to a viable architecture of getting exact representation of continuous, totally symmetric functions, which was proposed in the paper \textit{Deep Sets} by \citet{zaheer2018deepsets}. In this work, symmetric functions were interpreted as functions on sets because the order of the elements does not matter in a set. The concept of set-valued functions from this last work was extended to functions of multisets in \citet{xu2019powerful}. The main idea behind Deep sets is to process individual set elements in parallel using a shared encoding function and then combine them using a symmetric `pooling' function such as summation, average, or max-pooling. This idea behind deep sets was generalized considerably by the work on $k$-ary Janossy pooling by \citet{murphy2019janossy}. A very rigorous theoretical understanding of the latent dimensions of Deep Sets architecture and Janossy pooling paradigm was indicated in \citet{wagstaff2021universal}. The result of \citet{zaheer2018deepsets} was further corroborated by \citet{chen2023totsym} recently where they proved that any totally symmetric continuous function can be expressed as a composition of two continuous functions. To understand it better, we should introduce some terminologies in the following \citep{jegelka2022theorygraphneurnet}: 

\begin{definition}[Inner, Outer Functions, Embedding dimension]
    If a real valued, totally symmetric function evaluated at a set $S$ can be expressed as $f(S) =  \rho(\sum_{s \in S} \phi (s))$ where $\phi$ is independent of the $f$, then $\rho:\R^{d_1} \to \R$ is called the outer function and $\phi$ mapping to $\R^{d_1}$, is called the inner function. The dimension of the range of the inner function ($d_1$ here) is called embedding dimension. 
\end{definition}
The Deep sets ansatz states that a continuous totally symmetric function $f : \Omega^n \to \R$ for a compact $\Omega \in \R^d$, can be expressed as a composition of continuous inner and outer functions in the form $f (\bx)= \rho(\sum_{i=1}^n \phi (\bx_i))$. For $d=1$ the proof of this ansatz can be found in \cite{zaheer2018deepsets} and for $d>1$ it has been shown in \citet{chen2023totsym} that any continuous totally symmetric function $f$ can be written as $g \circ \bm{\eta}$ where $\bm{\eta} = (\eta_1, \ldots, \eta_M)$ is the collection of all generators of totally symmetric polynomials. Hence number of such generators is the embedding dimension here i.e. $M$. More about this can be found in Section \ref{sec:repCksympol}.  

There were not many results available about how to represent $n$-antisymmetric functions effectively, especially when $d>1$. There are some broad schemes of doing it, named as Backflow, Jastrow and Slater determinant ansatz \citep{zweig2023antisymmetric}. We now describe them briefly in the following. For $f, g$ complex valued functions from $ \Omega \subset \R^n$ we define $(f \otimes g): \Omega^2 \to \C$ as $(f \otimes g)(x,y)=f(x)g(y)$. With this notation one can define the antisymmetric projection of tensor product of functions as 
\begin{align*}
    \mathcal{A}(\phi_1 \otimes \ldots \otimes \phi_n)=\frac{1}{n!}\sum_{\sigma \in \calS_n} (-1)^{\sigma} \phi_{\sigma(1)} \otimes \ldots \otimes \phi_{\sigma(n)}.
\end{align*}
Up to some rescaling these projections are called the Slater determinant of the functions $\phi_1, \ldots \phi_n$. The functional form of Backflow ansatz (with a single term) is 
\begin{align*}
    p(\bx) \mathcal{A}(g_1 \otimes \ldots \otimes g_n)(\bx),
\end{align*}
where $p$ is a totally symmetric function. Similarly the functional form of Jastrow ansatz (with a single term) can be written as 
\begin{align*}
    \mathcal{A}(g_1 \otimes \ldots \otimes g_n)(\Phi(\bx)),
\end{align*}
where $\Phi(\bx):\mathbb{R}^n \to \R^n$ is an equivariant function i.e. $\Phi(\sigma. \bx)=\sigma. \Phi(\bx)$ (by $.$ we mean group action here). In the end the functional form of Slater determinant ansatz with $L$ terms is
\begin{align*}
    \sum_{\ell=1}^L \mathcal{A}(g_1^{\ell} \otimes \ldots \otimes g_n^{\ell})(\bx).
\end{align*}
It can be seen in \citet{zweig2023antisymmetric} that the Jastrow ansatz is special case of backflow ansatz and the Slater determinant ansatz is a special case of Jastrow. Some very interesting theoretical work around these ansatzs involving Slater determinants can be found in \cite{hutter2020antisym}, \cite{abrahamsen2023antisymmetric}.

In a very recent work the problem of exact representation of continuous $n$-antisymmetric functions has been solved in \citet{chenlu2023exactcontantisymmetric}. However we provide an example here that shows that their antisymmetric representation theorem cannot be directly used to give exact representations for $\calC^1$ or in general $\calC^k$ functions i.e. such a function $f$ cannot be written as $f=g \circ \bm{\eta}$ where $g$ has same regularity as $f$. Similar example in case of totally symmetric functions has been constructed in \citet[page~7]{chen2023totsym} where the function $f$ is $\calC^1$ yet $g$ cannot be so at a particular point. Before we present the counterexample, let us recall their recent work. According to \citet{chenlu2023exactcontantisymmetric}, given $d \geq 1$ and $n \geq 1$, a function $\bm{\eta}=(\eta_1, \eta_2, \ldots, \eta_m): (\R^d)^n \to \R^m$ satisfies \textit{assumption A} if 
\begin{enumerate}[label=(\roman*)]
    \item $\eta_k:(\R^d)^n \to \R$ is $n$-antisymmetric and continuous for each $k \in [n]$,
    \item $\bm{\eta}(\bx_1, \ldots, \bx_n)=\bm{0}$ if and only if $\bx_i=\bx_j$ for some $i,j \in [n]$ with $i \neq j$,
    \item If $\bm{\eta}(\bx_1, \ldots, \bx_n)=\bm{\eta}(\bx_1', \ldots, \bx_n') \neq \bm{0}$ then there exists a permutation $\sigma \in \calS_n$ such that $(\bx_1', \ldots, \bx_n')=(\bx_{\sigma(1)}, \ldots, \bx_{\sigma(n)})$.
\end{enumerate}
Then the following theorem was proved in their paper: 
\begin{theorem}\label{thm:chenluthm1antisym}
  Given $d,n \geq 1$ and $\Omega \in \R^d$ compact set, if $\bm{\eta}: \Omega^n \to \R^m$ satisfy \textit{assumption A} then for any $n$-antisymmetric, continuous function $f: \Omega^n \to \R$, there exists a unique $g:\bm{\eta}(\Omega^n) \to \R$ that is continuous and odd, satisfying  
  \begin{align*}
      f(\bx_1, \ldots, \bx_n)=g(\bm{\eta}(\bx_1, \ldots, \bx_n)), \ \text{for all} \ (\bx_1, \ldots, \bx_n) \in \Omega^n,
  \end{align*}
  where $\bm{\eta}(\Omega^n)$ is equipped with the topology induced from $\R^m$. 
\end{theorem}

To construct a counterexample of this theorem in the $\calC^1$ category, let us take $n=2$ and $d=1$. First we construct $\bm{\eta}=(\eta_1, \ldots, \eta_m)$ where $\eta_{i}: \mathbb{R}^2 \to \R$ satisfy \textit{assumption A} above. Let us take $\bm{\eta}=(\eta_1, \ldots, \eta_m)$ for $m \geq 4$ defined by $\eta_i(x_1,x_2)=(x_1^3-x_2^3)(x_1^{i-1}+x_2^{i-1})$, for $i \in [m]$. Then clearly $\eta_i$s are antisymmetric under the action of $\calS_2$, continuous and if $x_1=x_2$ then $\bm{\eta}(x_1,x_2)=\bm{0}$. Conversely, if $\bm{\eta}(x_1,x_2)=\bm{0}$ then it implies $x_1^3-x_2^3=0$ i.e. $x_1=x_2$. Also if $\bm{\eta}(x_1,x_2)=\bm{\eta}(y_1,y_2) \neq \bm{0}$ then from above we can note that $x_1 \neq x_2$ and $y_1 \neq y_2$. Then using the fact $m \geq 4$ and writing $\bm{\eta}$ explicitely, we get $x_1^3-x_2^3=y_1^3-y_2^3 \neq 0$ and $(x_1^3-x_2^3)(x_1^3+x_2^3)=(y_1^3-y_2^3)(y_1^3+y_2^3)$. This implies  $x_1^3-x_2^3=y_1^3-y_2^3 \neq 0$ and $x_1^3+x_2^3=y_1^3+y_2^3$ or in other words $x_1=y_1$ and $x_2=y_2$. Now let us take $f: \R^2 \to \R$ as $f(x_1,x_2)=x_1^{4/3}-x_2^{4/3}$. We see $f$ is continuous function that is antisymmetric and $\calC^1$ at the origin. Then according to Theorem \ref{thm:chenluthm1antisym} above there exists $g : \bm{\eta}(\R^2) \to \R$ that is continuous and odd and satisfies $f(\bx)=g(\bm{\eta}(\bx))$ for all $\bx \in \R^2$. However, now we show that this function $g$ cannot be $C^1$ (or even differentiable) at $\bm{\eta}(\bm{0})=\bm{0}$. Let us take $(x_1,x_2)=(\epsilon,0)=\bx(\epsilon)$. We investigate the differentiability of $g$ at $\bm{0}$ along $\bm{\eta}(\bx(\epsilon)) = (\epsilon^3, \epsilon^4 \ldots, \epsilon^{m+2})$ as $\epsilon \to 0$. We note $g(\bm{0})=0$ as $g$ is odd. If we have $g$ is differentiable at $\bm{0}$ with derivative of $g$ being $D_{g}(\bm{0})$ at $\bm{0}$, we must have   
\begin{align}\label{counterexlimit}
    \lim_{\epsilon \to 0} \frac{||g(\bm{\eta}(\bx(\epsilon)))-g(\bm{0})-D_{g}(\bm{0})\cdot \bm{\eta}(\bx(\epsilon))||}{||\bm{\eta}(\bx(\epsilon))||} = 0.
\end{align}
But $g(\bm{\eta}(\bx(\epsilon)))=f(\bx(\epsilon))=\epsilon^{4/3}$ by definition which gives us for small $\epsilon$, $||g(\bm{\eta}(\bx(\epsilon)))-g(\bm{0})-D_{g}(\bm{0})\cdot \bm{\eta}(\bx(\epsilon))||=||f(\bx(\epsilon))-D_{g}(\bm{0})\cdot \bm{\eta}(\bx(\epsilon))||=O(\epsilon^{4/3})$ whereas $||\bm{\eta}(\bx(\epsilon))||=O(\epsilon^{3})$ showing that the limit in equation \eqref{counterexlimit} does not exist.  

One should note that the exact representation theorems by Chen et al. above do not strictly fall under the Backflow ansatz mentioned earlier. Similarly \cite{han2022universalsymantisym} modified some results in the Deep Sets paper and showed that arbitrary uniform approximation of symmetric or antisymmetric functions defined on a compact subset of some Euclidean space is possible but the latent number of variables (embedding dimension) is dependent on the gradient of the function being approximated, and the order of approximation along with number $n$ and the dimension $d$, of the input variables. We state their exact results in the following: 
\begin{theorem}\label{Hantotallysymmetric}
    Let $f : \Omega^n \to \R$ be a continuously differentiable, totally symmetric functions, where $\Omega \subset \R^d$ is compact. If $0< \epsilon < ||\nabla f||_2\sqrt{nd} (n^{-1/d})$, then there exists $\phi: \R^d \to \R^M$, $g: \R^M \to R$ such that for any $\bx=(\bx_1, \ldots, \bx_n) \in \Omega^n$, we have 
    \begin{align*}
        \big|f(\bx) - g (\sum_{j=1}^n \phi(\bx_j))\big| \leq \epsilon, 
    \end{align*}
    where $M \leq 2^n(||\nabla f||_2^2nd)^{nd/2}/(\epsilon^nd n!)$ with $||\nabla f||_2=\max_{\bx}||\nabla f (\bx)||_2$. 
\end{theorem}

\begin{theorem}\label{Hanantisymmetric}
    Let $f : \Omega^n \to \R$ be a continuously differentiable, $n$-antisymmetric functions, where $\Omega \subset \R^d$ is compact. Then there exists $K$ permutation equivariant mappings $Y^k: (\R^d)^n \to \R^n$ and permutation invariant functions $U^k: (\R^d)^n \to \R$ for $k \in [K]$ such that for $\bx=(\bx_1, \ldots, \bx_n) \in \Omega^n$, we have 
    \begin{align*}
        \big|f(\bx) - \sum_{k=1}^K U^k(\bx) \prod_{i<j}(y_i^k(\bx)-y_j^k(\bx))\big| \leq \epsilon, 
    \end{align*}
    where $K \leq (||\nabla f||_2^2nd)^{nd/2}/(\epsilon^nd n!)$ and for each $U^k$ there exists $g^k:\R^d \to \R^m$, $\phi^k: \R^m \to \R$ with $m \leq 2^n$ such that for any $\bx=(\bx_1, \ldots, \bx_n) \in \Omega^n$,
    \begin{align*}
        U^k(\bx)=g^k(\sum_{j=1}^n \phi^k(\bx)).
    \end{align*}
\end{theorem}

\subsection{Our contributions}
\label{ssec:contrib}

In this paper, we prove results that can be grouped as two broad contributions:
\begin{enumerate}[label = (\Roman*) ]
    \item Let $\Omega \subset \R^d$ be compact and $G$ be a compact (Lie) group that acts on $[n]$. If $f : \Omega^n \to \R$ is $G$-invariant and $\calC^k$, then there exists a $G$-invariant polynomial $P$ that is arbitrarily close to $f$ under $\calC^k$-uniform norm on $\Omega^n$. This result remains true for permutation invariant $f$, i.e. $G = \calS_n$, which can be arbitrarily $\calC^k$ approximated by totally symmetric polynomials. Furthermore, since totally symmetric polynomials are finitely generated by totally symmetric power sums, its exact number of generators (embedding dimension) is $\binom{n+d}{d}$.
    \item Let $\Omega \subset \R^d$ be compact. If $f : \Omega^n \to \R$ is $n$-antisymmetric and $\calC^k$, then there exists an $n$-antisymmetric polynomial $P$ that is arbitrarily close to $f$ under $\calC^k$-uniform norm on $\Omega^n$. Furthermore, since $n$-antisymmetric polynomials form a finitely generated module over the totally symmetric polynomials, the exact number of module generators are stated in the following cases: When $n \geq 2$ and $d =1$, the number is 1; when $n = 2$ and $d \geq 1$, the number is $d$; when $n > 2$ and $d = 2$, the number is the $n$-th Catalan number $\frac{(2n)!}{(n+1)!n!}$ \citep{HaimanGarcia1996}. For the general case, the lower and upper bound for the minimum number of module generators are stated as $\dbinom{\binom{r}{d-1}}{j}$ and $\dbinom{\binom{n}{2}+dn}{dn}$, repectively, where $n = \binom{r}{d} + j$ for some $0 \leq j < \binom{r}{d-1}$.
\end{enumerate}

Essentially, the representation of the approximation in the totally symmetric case follows the same sum-decomposition of Deep Sets. Both the inner and outer functions can be chosen to be polynomials, and the inner function is chosen independently from the $f$. In addition, the embedding dimension is also independent of the $\epsilon$-accuracy of the approximation, the target function $f$, and the $\calC^k$-norm. These independencies are in contrast to Theorem \ref{Hantotallysymmetric} in the work of \citet{han2022universalsymantisym}; specifically, both the inner and outer functions can only be made to be continuous, and the inner function $\phi$ is dependent on $f$. The embedding dimension $M$ is dependent on the $\epsilon$-accuracy of the approximation, the target function $f$, and the gradient of $f$.

For the $n$-antisymmetric case, the representation of the approximation is a sum of $K$ terms, where each term is a product of a smooth totally symmetric polynomial and a smooth $n$-antisymmetric, homogeneous polynomial of degree at most $\binom{n}{2}$, which does not depend on the function $f$; in contrary, the functions $\{U^k\}_{k=1}^K$ in Theorem \ref{Hanantisymmetric} depends on the target function. Similarly, the number $K$ is also independent of the $\epsilon$-accuracy of the approximation, the target function $f$, and the gradient of $f$, unlike the one presented in \citet{han2022universalsymantisym}. The approximation results in these contributions are presented in Section \ref{sec:Cksymantisymapprox}, and the results for the representations are shown in Section \ref{sec:repofsymantisympol}.

\subsection{Structure of the paper}
\label{ssec:struc}
This paper is organized as follows: In Section \ref{sec:Cksymantisymapprox}, we prove the theorems of uniform, arbitrary $\calC^k$ approximations of $G$-invariant functions for some (Lie) group $G$, symmetric functions, and $n$-antisymmetric functions by such polynomials. Common notations are introduced in Section \ref{sec:notation}. Section \ref{sec:symGapprox} contains the necessary results and main proofs for uniform, $\calC^k$ approximating $G$-invariant and totally symmetric functions. A similar proof is given for $n$-antisymmetric functions in Section \ref{antisymGapprox}. 

Section \ref{sec:repofsymantisympol} is for the representations of totally symmetric and $n$-antisymmetric polynomials, which are used for uniform, $\calC^k$ approximations in Section \ref{sec:Cksymantisymapprox}. We discuss the embedding dimension needed for totally symmetric polynomials, which are generated as an $\R$-algebra following the work of \cite{chen2023totsym}. The $n$-antisymmetric polynomials are shown to form a finitely generated module over the totally symmetric functions in Section \ref{sec:repCkantisympol}, and the bounds for the minimal number of generators are also given. In the end, Table \ref{tab:summary_of_results} summarizes our results, which shows considerable improvement in representation of $n$-antisymmetric polynomials over existing literature.

Appendix \ref{sec:appendix} provides the necessary background on commutative algebra for the readers, which will be used in Appendix \ref{sec:appendix2}. Lastly, detail expositions on representation theory of general, finite $G$-invariant polynomials are given in Appendix \ref{sec:appendix2} along with necessary results on $n$-antisymmetric polynomials; these are used to infer on the minimal number of generators in Secton \ref{sec:repCkantisympol}.

\section{\texorpdfstring{\texorpdfstring{$\calC^k$}{Ck} approximation of $G$-invariant and $n$-antisymmetric functions}{}}
\label{sec:Cksymantisymapprox}

The approximation theory presented in this section applies to polynomials with coefficients in both $\R$ or $\C$. We will focus on the proof for real polynomials here. For complex-valued polynomials, the same results are true, and we simply need to consider our results for the real and imaginary parts of the polynomial separately.  


\subsection{Notation}
\label{sec:notation}

The $\calC^k$ approximations for $G$-invariant functions will be shown for a compact Lie group $G$, which has an action on $[n] = \{1, 2, \ldots, n\} $; the results for totally symmetric and $n$-antisymmetric functions will then actually follow quite closely from the proof of $G$-invariant functions. Notably, given such a compact Lie group $G$, $\sigma \in G$ is a bijection on $[n]$, so the action of $\sigma$ is simply a permutation in $\calS_n$. Therefore, one of the essences of proving our theorems is understanding how the permutations behave on the indices of $\bx = (\bx_1, \ldots, \bx_n) \in (\R^d)^n$ and their involvement in the differentiation for the $\calC^k$ approximation. The other essence is to consider an appropriate function space that will eventually guarantee the existence of the $\calC^k$ approximation, which we will describe next.

The following $\R$-algebra (see Definition~\ref{def:algebra}) and topology are needed to describe a subalgebra that will be crucial for our initial $\calC^k$ approximation of these functions. Suppose $U \subset (\R^d)^n$ is an open set. Define $\calC^k(U; \R)$ as the $\R$-algebra of all $k$ times continuously differentiable real valued functions on $U$, and it is abbreviated to $\calC^k(U)$, as the codomain of these functions will always be $\R$. This space is endowed with the topology $\tau_u^k$, which is the compact-open topology of order $k$; in other words, it is the topology of uniform convergence for the functions and all their partial derivatives up to the $k^{\text{th}}$-order on compact subsets of $U$. Next, notice that if $V \subset (\R^d)^n$ is compact, then there always exists an open set $U$ such that $V \subset U \subseteq (\R^d)^n$. Pick any such $U$, and then we may define the vector space $\calC^k(V) := \{f|_{V}: f \in \calC^k(U)\}$
of $k$-times continuously differentiable functions on $V$, which also forms an $\R$-algebra. We will equip this space with the $\calC^k$-uniform norm defined as the maxima over continuous derivatives up to order $k$, restricted to the compact set $V$: 
\begin{definition}[$\calC^k$ norm]
\label{def:Ck_norm}
    Let $V \subset (\R^d)^n$ be compact, and suppose $p \in \calC^k(V)$. Then we define
    \begin{equation}
    \label{eq:Ck_norm}
    \| p \|_{\calC^k(V)}
        \coloneqq 
        \sum_{|\alpha| \leq k} \max_{\bx \in V} \left| (D^\alpha p) (\bx) \right|,
    \end{equation}
    where $\alpha \coloneqq (\alpha_1, \ldots, \alpha_{dn}) \in \ZZ^{dn}$ is a $dn$-dimensional multi-index, $|\alpha| := \alpha_1 + \ldots + \alpha_{dn}$, and
    \begin{equation}
        D^\alpha p := \frac{\partial^{\alpha_1}}{\partial x_1^{\alpha_1}} \cdots \frac{\partial^{\alpha_{dn}}}{\partial x_{dn}^{\alpha_{dn}}} p.
    \end{equation}
\end{definition}
With the norm $\lVert \cdot \rVert_{\calC^k(V)}$, the vector space $\calC^k(V)$ turns into a Banach space, and is independent of the choice of the open set $U$.

Let us introduce some notations to describe the permutations and multi-indices that are involved for the general case $d \geq 1$. Recall that for $d = 1$, $\bx \in \Omega^n$ may be written as $\bx = x_1\be_1 + \ldots + x_n\be_n$ using the standard basis. Then for $d \geq 1$, $\bx \in \Omega^n$ may be viewed as a $dn$-dimensional vector, which can be written similarly. In particular, a basis $\{\be_\ell \}_{\ell = 1}^{dn}$ is needed, but for practicality, we introduce
\begin{equation}
\label{eq:eij}
    \be_{i,j} = \be_{d(i-1) + j},
\end{equation}
where $i \in [n]$ and $j \in [d]$. Hence for any $\bx = (\bx_1, \ldots, \bx_n) \in \Omega^n$, we may write $\bx = \sum_{i=1,j=1}^{n,d} x_{ij}\be_{i,j}$. This representation for the general case is relevant below in showing that the action $\sigma.\bx$ can be described as an action on the $i$-indices of the basis $\{ \be_{i,j} \}_{i=1,j=1}^{n,d}$, instead of the indices of the vector $(\bx_1, \ldots, \bx_n)$.


\subsection{\texorpdfstring{$\calC^k$}{Ck} approximation via \texorpdfstring{$G$}{G}-invariant and totally symmetric polynomials}
\label{sec:symGapprox}

In this section, we are interested in the $\calC^k$ approximation of totally symmetric functions and, in general, $G$-invariant functions for some compact (Lie) group $G$ which has an action on $[n]$. We first attain new observations of permutations on the indices of $\bx \in \Omega^n$ and then see the behaviors affected by differentiation. Subsequently, proofs for the $\calC^k$ approximations are given.

The main theorem of our section is the approximation for $G$-invariant $f \in \calC^k$ where $G$ is a compact (Lie) group. This result certainly will cover the case of $G = \calS_n$ because it is a finite group. We have the main theorem and its corollary:

\begin{theorem}[$\calC^k$ approximation: $G$-invariant function]
\label{thm:G_invariant_polynomial}
    Let $\Omega \subset \R^d$ be compact, and assume $G$ is a compact (Lie) group that acts on $[n]$. Let $f: \Omega^ n\to \R$ be $G$-invariant and $\calC^k$. Then for every $\epsilon > 0$, there exists a $G$-invariant polynomial $P$ such that 
    \begin{equation}
    \label{eq:G_invariant_polynomial}
        \left \| f-P \right \|_{\calC^k} < \epsilon.
    \end{equation}
\end{theorem}

\begin{corollary}[$\calC^k$ approximation: totally symmetric function]
\label{cor:Ck_approximation_symmetric_polynomial}
   Suppose $\Omega \subset \R^d$ is compact. Let $f: \Omega^n \to \R$ be totally symmetric and $\calC^k$. Then for every $\epsilon > 0$ there exists a totally symmetric polynomial $P$ such that 
    \begin{equation}
        \left\| f-P \right\|_{\calC^k} < \epsilon.
    \end{equation}
\end{corollary}

Note that for any group $G$ acting on $[n]$, $\Omega^n$ is $G$-invariant, and this is true by constructing $\Omega^n$ to be the $n$-product of $\Omega$. In fact, the assumptions in the results above (and later results) can be looser. The approximation results are true for 
some general compact set $K \subset (\R^d)^n$ because $K$ can always be made  $G$-invariant by the following argument: suppose $G$ is a compact (Lie) group that acts on $[n]$; then for any $\sigma \in G$, $\sigma$ is a bijection on $[n]$, so
    \[
    K' = \bigcup_{\sigma \in G} \sigma(K) = \bigcup_{\sigma \in H \subseteq \calS_n} \sigma(K),
    \]
for some subgroup $H$ of $\calS_n$. Note that $\sigma(K)$ is compact since $\sigma$ is continuous on the compact set $K$, and since $\left| H \right| \leq \left| \calS_n \right| = n!$ is finite, $K'$ is compact as it is the finite union of compact sets. Now, it is clear that $K'$ is $G$-invariant, so it may replace $\Omega^n$ in the assumptions of our results.

Now, our proof actually begins with Nachbin's characterization \citep{nachbin1949algebres} that the set of polynomials of real coefficients with $dn$-variables (a subalgebra) is $\tau_u^k$-dense in $\calC^k$. The next theorem is a consequence of Nachbin's Theorem. 

\begin{theorem}[Theorem 2.2, \cite{Prolla_extensionnachbin76}]
\label{thm:Nachbins}
    Suppose $U \subset \R^m$ is open and $\calC^k(U)$ is endowed with the topology $\tau_u^k$. If $\calA \subset \calC^k(U)$ be a polynomial algebra, then $\calA$ is $\tau_u^k$-dense in $\calC^k(U)$ if and only if the following conditions are satisfied
    \begin{enumerate}[label = (\alph*)]
        \item For any $x,y \in U$ with $x \neq y$, there exists $f \in \calA$ such that $f(x) \neq f(y)$.
        \item For any $x \in U$, there exists $f \in \calA$ such that $f(x) \neq 0$. 
        \item For any $x \in U$ and $u \in \R^m$ with $u \neq 0$, there exists $f \in \calA$ such that $D_u f(x) \neq 0$. 
    \end{enumerate}
\end{theorem}

It is easy to see that our subalgebra, the set of polynomials with real coefficient is $\tau_u^k$-dense in the set $\calC^k$, so Theorem \ref{thm:Nachbins} will be integral in our proof later. Now, we look to observe how $\sigma$ acts on $\bx$ and behave in the differentiation for some $\sigma \in \calS_n$:

\begin{lemma}
\label{lemma:sigma_action}
    Suppose $d\geq1$ and $\bx \in \Omega^n$. Let $\sigma \in \calS_{n}$, then 
    \begin{equation}
    \label{eq:sigma_action}
        \sigma.\bx = \sum_{i=1,j=1}^{n,d} x_{ij} \be_{\sigma^{-1}(i),j}
    \end{equation}
\end{lemma}
\begin{proof}
    It suffices to show this for transpositions in $\calS_n$. Without loss of generality, let $\sigma = (1 \; 2)$.
    \begin{align*}
        &\sigma.\bx
        = \sum_{j=1}^d x_{\sigma(1)j}\be_{1,j}  + x_{\sigma(2)j}\be_{2,j} + \ldots +  x_{\sigma(n)j}\be_{n,j}
        = \sum_{j=1}^d x_{2j}\be_{1,j} + x_{1j}\be_{2,j} + \ldots + x_{nj}\be_{n,j}\\
        &\hspace{-1mm}=\sum_{j=1}^d x_{1j}\be_{2,j} + x_{2j}\be_{1,j} + \ldots + x_{nj}\be_{n,j} 
        = \sum_{j=1}^d x_{1j}\be_{\sigma^{-1}(1),j} + x_{2j}\be_{\sigma^{-1}(2),j} + \ldots + x_{nj}\be_{\sigma^{-1}(n),j}.
    \end{align*}
    This completes our proof.
\end{proof}

The next lemma will be important for us to bridge that connection between $D^\alpha (P \circ \sigma)(\bx)$ and $D^\alpha P(\bx)$ in the proof of Theorem \ref{thm:G_invariant_polynomial}. A general version of this lemma can be found in \citet[page~261]{Kaneinvarianttheorybook}, but we present a more elementary proof for the easier case here. Even though the lemma is proven for $\calS_n$, this can be applied for any group $G$ that has an action on $[n]$ because any such element $g \in G$ induces a permutation of $[n]$.  
\begin{lemma}
\label{lem:Dalpha}
Suppose $p : U \subseteq (\R^d)^n \to \R$ is smooth. Let $\alpha \in \ZZ^{dn}$ be a multi-index and $\sigma \in \calS_n$, then 
    \begin{equation}
    \label{eq:Dalpha}
        D^\alpha(p \circ \sigma)(\bx) =D^{\sigma. \alpha} p (\sigma(\bx))
    \end{equation}
\end{lemma}
\begin{proof}
We prove this for the general space dimension $d \geq 1$, and we will prove this statement using induction on the order of $|\alpha|$. For base case $|\alpha| = 0$, we have $D^\alpha p = p$, so the statement above is true trivially. For base case $|\alpha| = 1$, we take $\alpha = \be_{i,j} = \be_{d(i-1) +j} \in \ZZ^{dn}$ for $1 \leq i \leq n$ and $1 \leq j \leq d$. We compute $\frac{\partial}{\partial y_{ij}}(p \circ \sigma)(\by) \lvert_{\by = \bx}$, which involves the chain rule and the term $\frac{\partial \sigma}{\partial y_{ij}}(\by)$. Hence, let us compute this first:
    \begin{align*}
    \frac{\partial \sigma}{\partial y_{ij}} (\by)
    &= \frac{\partial}{\partial y_{ij}} \left( \sum_{\ell = 1}^d y_{\sigma(1)\ell} \be_{1,\ell} + \ldots + y_{\sigma(i)\ell} \be_{i,\ell} + \ldots y_{\sigma(n)\ell} \be_{n,\ell}  \right)\\
    &= \frac{\partial}{\partial y_{ij}} \left( \sum_{\ell = 1}^d y_{1\ell} \be_{\sigma^{-1}(1),\ell} + \ldots + y_{i\ell} \be_{\sigma^{-1}(i),\ell} + \ldots + y_{n\ell} \be_{\sigma^{-1}(n),\ell} \right)
    = \be_{\sigma^{-1}(i),j},
    \end{align*}
where the second equality follows from lemma~\ref{lemma:sigma_action}. Now, we write
    \begin{align*}
        &D^\alpha(p \circ \sigma)(\by) \bigg \lvert_{\by = \bx} 
        =\frac{\partial}{\partial y_{ij}}(p \circ \sigma)(\by) \bigg \lvert_{\by = \bx} 
        = \nabla_{\by} p(\sigma(\by)) \cdot \frac{\partial \sigma}{\partial y_{ij}}(\by) \bigg \lvert_{\by = \bx} \\
        &= \left(\frac{\partial p}{\partial y_{11}}(\sigma(\by)), \ldots, \frac{\partial p}{\partial y_{nd}}(\sigma(\by))  \right) \cdot \be_{\sigma^{-1}(i),j} \bigg \lvert_{\by = \bx} 
        = \frac{\partial p}{\partial y_{\sigma^{-1}(i)j}}(\sigma(\by)) \bigg \lvert_{\by = \bx} 
        = D^{\sigma . \alpha}p(\sigma(\bx)),
    \end{align*}
since $\sigma.\alpha = \sigma(\be_{i,j}) = \be_{\sigma(-1),j}$. This concludes the base cases.

Assume that the statement is true for $|\alpha| = k$ and we want to show that it is also true for $|\beta| = k + 1$. Let $\beta = \alpha + \be_{i,j}$, then 
    \begin{align*}
        D^\beta(p \circ \sigma)(\by) \bigg \lvert_{\by = \bx}
        &= \frac{\partial}{\partial y_{ij}}(D^\alpha(p \circ \sigma))(\by) \bigg \lvert_{\by = \bx} 
        = \frac{\partial}{\partial y_{ij}}(D^{\sigma.\alpha}p )(\sigma(\by)) \bigg \lvert_{\by = \bx} \\
        &= \nabla_{\by} (D^{\sigma.\alpha}p )(\sigma(\by)) \cdot \be_{\sigma^{-1}(i),j} \bigg \lvert_{\by = \bx} 
        = \frac{\partial D^{\sigma.\alpha}p}{\partial y_{\sigma^{-1}(i)j}}(\sigma(\by))\bigg \lvert_{\by = \bx} 
        = D^{\sigma. \beta}p(\sigma(\bx))
    \end{align*}
because $\sigma. \beta = \sigma.(\alpha + \be_{i,j}) = \sigma.\alpha + \be_{\sigma^{-1}(i),j}$. The second equality follows from the induction hypothesis.
\end{proof}

\begin{proof}{of Theorem \ref{thm:G_invariant_polynomial}.}
    
    First, we approximate $f \in \calC^k$ up to the $k$-order derivatives by a polynomial $\hat{P}$ using Theorem \ref{thm:Nachbins}. This allows us to assume that 
    \begin{equation}
    \label{eq:G_inv_proof_eq1}
        \| f-\hat{P} \|_{\calC^k} = \sum_{|\alpha| \leq k} \max_{\bx \in \Omega^n} | D^\alpha ( f-\hat{P} ) |(\bx) < \epsilon.
    \end{equation}

    Now consider the symmetrized polynomial $P$ defined by
    \[
        P(\bx) = \int_{G} (\hat{P} \circ \sigma) (\bx) \, d\mu(\sigma),
    \]
    where $d\mu(\sigma)$ is the Haar probability measure associated with the group $G$. Similarly, since $f$ is $G$-invariant, it can be written in the same form: $f(\bx) = \int_{G} \left(f \circ \sigma \right) (\bx) \, d\mu(\sigma)$.

    \begin{equation*}
        \begin{split}
        &\left \| f-P \right \|_{\calC^k} 
        = \sum_{|\alpha| \leq k} \max_{\bx \in \Omega^n} \left| D^\alpha\left( f-P \right) \right|(\bx) 
        \overset{(a)}{=} \sum_{|\alpha| \leq k} \max_{\bx \in \Omega^n} \left| D^\alpha \int_G (( f-\hat{P}) \circ \sigma)(\bx) \, d \mu(\sigma)\right| \\
        &\overset{(b)}{=} \sum_{|\alpha| \leq k} \max_{\bx \in \Omega^n} \left| \int_G D^\alpha (( f-\hat{P}) \circ \sigma)(\bx) \, d \mu(\sigma)\right| 
        \overset{(c)}{=} \sum_{|\alpha| \leq k} \max_{\bx \in \Omega^n} \left| \int_G D^{\sigma.\alpha} (f-\hat{P})(\sigma(\bx)) \, d \mu(\sigma)\right|  \\
        &\leq \sum_{|\alpha| \leq k} \max_{\bx \in \Omega^n} \int_G |D^{\sigma.\alpha} (f-\hat{P})| (\sigma(\bx)) \, d \mu(\sigma) 
        \overset{(d)}{\leq} \sum_{|\alpha| \leq k} \int_G \max_{\bx \in \Omega^n}  |D^{\sigma.\alpha} (f-\hat{P})| (\sigma(\bx)) \, d \mu(\sigma)  \\
        &\overset{(e)}{=} \sum_{|\alpha| \leq k} \int_G \max_{\bx \in \Omega^n}  |D^{\sigma.\alpha} (f-\hat{P})| (\bx) \, d \mu(\sigma) 
        =\int_G  \sum_{|\alpha| \leq k}  \max_{\bx \in \Omega^n}  |D^{\sigma.\alpha} (f-\hat{P})| (\bx) \, d \mu(\sigma) \\ 
        &\overset{(f)}{=}\int_G  \sum_{|\alpha| \leq k}  \max_{\bx \in \Omega^n}  |D^{\alpha} (f-\hat{P})| (\bx) \, d \mu(\sigma)
        < \epsilon.
        \end{split}
    \end{equation*}

In the above equation, $(a)$ follows by symmetrization, and $(c)$ follows from Lemma \ref{lem:Dalpha}. We now briefly explain the reasoning for the other steps. $(b)$ follows by noting that for any $|\alpha| \leq k$, $| D^\alpha (f-\hat{P})|$ is always bounded on $\Omega^n$ because $\Omega^n$ is compact. Hence, letting $| D^\alpha (f-\hat{P})| \leq M $ for all $\alpha$, and since $M \in L^1(\mu)$, the differentiation can be passed through the integral. To obtain $(d)$, we note that $|D^{\sigma.\alpha} (f-\hat{P})| (\sigma(\bx)) \leq \max_{\bx \in \Omega^n} |D^{\sigma.\alpha} (f-\hat{P})| (\sigma(\bx))$ for any $\sigma \in G$, and thus integrating over the group retains the inequality
    \[
    \int_G |D^{\sigma.\alpha} (f-\hat{P})| (\sigma(\bx)) \, d\mu(\sigma) \leq \int_G \max_{\bx \in \Omega^n} |D^{\sigma.\alpha} (f-\hat{P})| (\sigma(\bx)) \, d\mu(\sigma).
    \]
$(e)$ follows because we are taking the maximum. $(f)$ follows because for a fixed $\sigma \in G$ and all the $\alpha$ such that $|\alpha| = l$, $\sigma$ is a bijection on the list of $\alpha$'s, so it is the same list of $\alpha$'s. 

\end{proof}

\begin{remark}
\label{rem:corollary}
    The proof of Corollary \ref{cor:Ck_approximation_symmetric_polynomial} follows the proof for Theorem \ref{thm:G_invariant_polynomial} closely with a small difference. Given $G = \calS_n$, $G$ is now finite and discrete. In this case, the Haar measure becomes a normalized counting measure where we replace integration by a summation with a normalization factor $\frac{1}{n!}$ because $|\calS_n| = n!$. 
\end{remark}

\subsection{\texorpdfstring{$\calC^k$}{Ck} approximation via \texorpdfstring{$n$}{n}-antisymmetric polynomials}
\label{antisymGapprox}

In this section, we prove a similar $\calC^k$ approximation for $n$-antisymmetric functions. The formalism in this section will mirror that which was presented in Section \ref{sec:symGapprox}. There is a notion of `skew-symmetric' polynomials that generalize the notion of antisymmetry but in that the group $G$ has to be generated by reflections or pseudo-reflections. For such compact groups $G$ acting on $[n]$, our results below will hold true. However, we will only work with $\calS_n$ in our study here. Interested readers may find the notion of skew-symmetric polynomials in \citet[page~93]{Bergalgcombcoinvspaces}.  

\begin{theorem}[$\calC^k$ approximation: $n$-antisymmetric function]
\label{thm:Ck_approximation_antisymmetric_polynomial}
    Let $\Omega \subset \R^d$ be compact. Suppose that $f: \Omega^n \to \R$ is $n$-antisymmetric and $\calC^k$. Then for every $\epsilon> 0$ there exists an $n$-antisymmetric polynomial $P$ such that 
    \begin{equation}
        \left\| f-P \right\|_{\calC^k} < \epsilon.
    \end{equation}
\end{theorem}
\begin{proof}
    Using Theorem \ref{thm:Nachbins}, we approximate $f \in \calC^k$ up to the $k$-order derivatives by a polynomial $\hat{P}$. Hence, let 
    \begin{equation}
    \label{eq:antsym_proof_eq1}
        \| f-\hat{P} \|_{\calC^k} = \sum_{|\alpha| \leq k} \max_{\bx \in \Omega^n} | D^\alpha ( f-\hat{P} ) |(\bx) < \epsilon.
    \end{equation}
    
    Now consider the $n$-antisymmetrized polynomial $P$ defined by
    \[
        P(\bx) = \frac{1}{n!}\sum_{\sigma \in \calS_n} \sgn(\sigma) (\hat{P} \circ \sigma) (\bx).
    \]
    Similarly, $f(\bx) = \sum_{\sigma \in \calS_n} \sgn(\sigma) \left(f \circ \sigma \right) (\bx)$ since $f$ is $n$-antisymmetric.
    
    We prove the approximation in a similar manner as in the proof of Theorem~\ref{thm:G_invariant_polynomial}.
    
    \begin{align*}
        &\left \| f-P \right \|_{\calC^k} 
        = \sum_{|\alpha| \leq k} \max_{\bx \in \Omega^n} \left| D^\alpha\left( f-P \right) \right|(\bx) \\
        &= \sum_{|\alpha| \leq k} \max_{\bx \in \Omega^n} \left|  D^\alpha \left(\frac{1}{n!} \sum_{\sigma \in \calS_n} \sgn(\sigma) (( f-\hat{P}) \circ \sigma)(\bx) \right) \, \right| \\
        &= \sum_{|\alpha| \leq k} \max_{\bx \in \Omega^n} \left| \frac{1}{n!}\sum_{\sigma \in \calS_n} \sgn(\sigma) D^\alpha (( f-\hat{P}) \circ \sigma)(\bx) \, \right| \\
        &= \sum_{|\alpha| \leq k} \max_{\bx \in \Omega^n} \left| \frac{1}{n!}\sum_{\sigma \in \calS_n} \sgn(\sigma) D^{\sigma.\alpha} (f-\hat{P})(\sigma(\bx)) \, \right| \\
        &\leq \sum_{|\alpha| \leq k} \max_{\bx \in \Omega^n} \frac{1}{n!}\sum_{\sigma \in \calS_n} |D^{\sigma.\alpha} (f-\hat{P})| (\sigma(\bx)) \,   
        \leq \sum_{|\alpha| \leq k} \frac{1}{n!}\sum_{\sigma \in \calS_n}  \max_{\bx \in \Omega^n}  |D^{\sigma.\alpha} (f-\hat{P})| (\sigma(\bx)) \,  \\
        &= \sum_{|\alpha| \leq k} \frac{1}{n!}\sum_{\sigma \in \calS_n}  \max_{\bx \in \Omega^n}  |D^{\sigma.\alpha} (f-\hat{P})| (\bx) \,  
        =\frac{1}{n!}\sum_{\sigma \in \calS_n} \sum_{|\alpha| \leq k}  \max_{\bx \in \Omega^n}  |D^{\sigma.\alpha} (f-\hat{P})| (\bx) \, \\
        &=\frac{1}{n!}\sum_{\sigma \in \calS_n}  \sum_{|\alpha| \leq k}  \max_{\bx \in \Omega^n}  |D^{\alpha} (f-\hat{P})| (\bx) \  < \epsilon .
    \end{align*}
   
\end{proof}
\section{\texorpdfstring{Representations of totally symmetric and $n$-antisymmetric polynomials}{}}
\label{sec:repofsymantisympol}
The representation theory presented in this section applies to polynomials with coefficients in any field of characteristics zero. However, we will mainly show the results for real polynomials. 
\subsection{Representations of the totally symmetric polynomials}
\label{sec:repCksympol}

In Section \ref{sec:symGapprox}, a totally symmetric $f \in \calC^k$ is shown to be uniformly approximated by a totally symmetric polynomial $P$ over $\Omega^n$, where $\Omega \subset \R^d$ is compact. In this section, we infer more about this totally symmetric polynomial $P$. 

For consistency, we are using the notations in \citet{chen2023totsym} which reference from \citet{emmbrian_whenmltsympoly} (page 359, Theorem 3). Let us denote $\mathcal{P}^{d,n}_{\text{sum}}(\R)$ as the $\R$-algebra consisting of all multi-symmetric polynomials with real coefficients, i.e. real totally symmetric polynomials in $\bx = (\bx_1,\ldots, \bx_n) \in (\R^d)^n$. Furthermore, $\mathcal{P}^{d,n}_{\text{sum}}(\R)$ is generated as an $\R$-algebra by multi-symmetric power sums:
\begin{equation} 
\label{eq:etarephs}
    \begin{aligned}
            \eta_{\bs}(\bx_1,\ldots, \bx_n) & \coloneqq \sum_{i=1}^n x_{i 1}^{s_1} x_{i 2}^{s_2} \cdots x_{i d}^{s_d}, \ \ 0 \leq s_1 + s_2 + \cdots + s_d \leq n, \\
    &= \sum_{i=1}^n h_{\bs}(\bx_i)
    \end{aligned}
\end{equation}
where $\bs= (s_1, s_2, \ldots, s_d)$ and $\bx_i \coloneqq (x_{i1}, \ldots, x_{i d})$. One can define a total ordering on these $\bs$ indices and enumerate them as $\eta_1, \eta_2, \ldots, \eta_M$, with $\eta_{j}(\bx_1,\ldots, \bx_n) = \sum_{i=1}^n h_{j}(\bx_i)$ from above for all $j \in [M] = \{1, 2, \ldots, M\} $. One can easily check that 
\begin{equation}
    M 
    = \sum_{i=0}^n \binom{i+d-1}{d-1} 
    = \sum_{i=0}^n \binom{i+d-1}{i} 
    = \binom{n+1+d-1}{n} 
    = \binom{n+d}{n}.
\label{eqn:possible_sums}
\end{equation}

Then $P$ can be written using the generators of the algebra: 
\begin{align}
\label{eq:grepgen}
    P = \sum_{k=1}^p C_k \prod_{i=1}^M\eta_{i}^{k,i},
\end{align}
where $C_k \in \R$. Now, based on the structures of $\{\eta_j \}_{j=1}^M$ and $P$, we can further simplify our representation. First, encode the information of $\{\eta_j \}_{j=1}^M$ in an $(M \times n)$ matrix $\Theta$, whose entries are 
\begin{equation}
\label{eq:bigmatrixallinfoMn}
    (\Theta)_{k \ell} = h_k(\bx_{\ell})
\end{equation}
for $k \in [M]$ and $\ell \in [n]$. Also, note that the $\ell$-th column of the matrix $\Theta$ can be written as  
\begin{align*}
    \Theta_{\ell} 
    = \phi(\bx_{\ell}) 
    \coloneqq (h_1 (\bx_{\ell}), h_2 (\bx_{\ell}),\ldots, h_M (\bx_{\ell}))^T.
\end{align*}

By \eqref{eq:grepgen}, $P$ is a function in terms of $\{\eta_i\}_{i=1}^M$; then from \eqref{eq:etarephs}, $P$ is a function in terms of $\sum_{i=1}^n \phi(\bx_{i})$, where the sum is the so-called inner function. This functional dependence is clearly smooth. Dimension of the range of this function $\phi$ is the `embedding dimension', which is $M = \binom{n+d}{n}$. This is a considerable improvement compared to \citet{han2022universalsymantisym}, where the upper bound on the embedding dimension depends on $\epsilon$ and the norm of the gradient of the approximated function.

\subsection{Representations of the \texorpdfstring{$n$}{n}-antisymmetric polynomials}
\label{sec:repCkantisympol}

This section now utilizes a fair amount of commutative algebra. The relevant definitions and results for this section can be found in Appendices \ref{sec:appendix} and \ref{sec:appendix2}. Given the polynomial ring $\R[\bx_1, \ldots, \bx_n]$ where $\{\bx_i\}_{i=1}^n \subset \R^d$, the representation of $n$-antisymmetric polynomials is a more subtle problem than that of totally symmetric ones. Unlike symmetric polynomials, $n$-antisymmetric polynomials do not form an $\R$-algebra because the closure axiom fails in general. Hence, we cannot talk about the generators for this algebra. However, we shall find representations for $n$-antisymmetric polynomials with respect to the totally symmetric polynomials, where the representation of the latter has been discussed in Section \ref{sec:repCksympol}.

\subsubsection{Finite generation of \texorpdfstring{$n$}{n}-antisymmetric polynomials and the case \texorpdfstring{$n=2$}{n2}}

In fact, the $n$-antisymmetric polynomials form a finitely generated module over the ring of totally symmetric polynomials (Appendix \ref{sec:appendix2}, Lemma \ref{lem:finitegenatisymonsym} ). Consequently, we want to determine the minimum number of module generators of $n$-antisymmetric polynomials over the totally symmetric polynomial ring; this is a required analysis because unlike a vector space over a field, there is not a single number of generators for a finitely generated module. The solution for the exact minimum number of module generators for the general case of $n \geq 2$ and $d \geq 1$ is daunting and very difficult to determine, so instead we establish some upper bounds (and lower bounds) on this minimality. This is the main result of the section, which is discussed in Section \ref{sec:bounds_on_generators_anti}. In addition, we also establish the minimality for specific cases here as well.

In the case $n \geq 2$ and $d = 1$, the minimum number of generators is 1, which is well known, and goes back to \citet{Cauchybookwherevandermondeisthere}. In particular, any $n$-antisymmetric polynomial can be written as a product of a symmetric polynomial and Vandermonde determinant $D(\bx) = \prod_{1 \leq i<j \leq n}(x_i-x_j)$. In other words, if $f$ is $n$-antisymmetric then there exists a symmetric polynomial $g$ such that $f(\bx) = D(\bx)g(\bx)$. Since for any $\sigma \in \calS_n$, $D(\sigma.\bx) = \sgn(\sigma) D(\bx)$ and $g(\sigma.\bx) = g(\bx)$, the result of the action is $f(\sigma.\bx) = \sgn(\sigma)f(\bx)$. We make a useful observation that for any transposition $(i \; j) = \sigma \in \calS_n$ where $i \neq j$, $f(\sigma.\bx)$= - $f(\bx)$ implies that $f$ vanishes on the hyperplane $x_i = x_j$; this means $f$ is divisible by $(x_i - x_j)$.

For $n = 2$ and $d > 1$, the solution takes the inspiration from the observation above, which will be used in its proof. First, let us consider Vandermonde determinants of all the $n$ scalar variables coming from $\bx_1, \ldots, \bx_n \in \R^d$; in other words, let $D_{l} (\bx) = \prod_{1 \leq i<j \leq n}(x_{i l}-x_{j l})$ where $l \in \{1, 2, \ldots, d\}$. The result is proven in the following lemma:

\begin{lemma}
\label{lem:antisymetric_n_eq_2}
    Let $n = 2$ and $d \geq 1$. Any $n$-antisymmetric polynomial $f$ can be written in the following form: 
    \begin{align}\label{eq: antisympolrepdg1}
        f(\bx) = \sum_{l=1}^d D_{l}(\bx) g_l(\bx),
    \end{align}
    where $g_1, \ldots, g_d$ are totally symmetric polynomials.
\end{lemma}

\begin{proof}
    In this particular case, $D_{l} (\bx) = (x_{1 l}-x_{2 l})$ where $l \in \{1, \ldots, d\}$, and in this proof, we write $\bx = (\bx_1, \bx_2) \in (\R^d)^2$. Let $\sigma$ be the permutation transforming $\bx \mapsto (\bx_2,\bx_1)$, so the $2$-antisymmetry of $f$ can be written as $-f(\bx) = -f(\bx_1, \bx_2) = f(\bx_2, \bx_1) = f(\sigma.\bx)$. The $\sigma$-transformation on $\bx$ is identified with the matrix  
    \begin{align}
       \Pi= \begin{pmatrix}
            0_d & I_d \\
            I_d & 0_d 
        \end{pmatrix},
    \end{align}
    where $I_d$ is the $d \times d$ identity matrix and $0_d$ is the $d \times d$ zero matrix.
    To arrive at the proposed form, we observe the 2-antisymmetry in a different coordinate via a linear transformation and factor out $D_l(\bx)$ terms. First, let $y_{1i}=x_{1i}-x_{2i}$ and $y_{2i}=x_{1i}+x_{2i}$ for $i \in \{1, \ldots, d\}$. This matrix of transformation is
    \begin{align}
        L = \begin{pmatrix}
            I_d & -I_d \\
            I_d & I_d
        \end{pmatrix},
    \end{align}
    so that $\by = (\by_1, \by_2) = L (\bx_1, \bx_2) = L \bx$. If $P = f \circ L^{-1}$, which is also a polynomial,
    the condition $-f(\bx) = f(\sigma. \bx)$ translates in the new coordinates as $-P(\by)=P(L \Pi L^{-1} \by)$. One can easily compute $L \Pi L^{-1}$ and see 
    \begin{align}
    \label{eq:antisymmterynewcoord}
        -P(\by_1,\by_2)=P(-\by_1, \by_2). 
    \end{align}
    Now, if $P$ is expressed as an $\R$-linear combination of monomials, then this last condition 
    is also valid for each monomial term. We see this by writing $P(\by) = P_1(\by) + P_2(\by)$ where 
    \begin{align*}
        P_1(\by_1,\by_2) \coloneqq \sum_{\alpha} c_{\alpha} m_{\alpha}(\by_1,\by_2) \quad\mathrm{and}\quad P_2(\by_1,\by_2) \coloneqq \sum_{\alpha} d_{\alpha} n_{\alpha}(\by_1,\by_2).
    \end{align*}
    Furthermore, $m_{\alpha}$ and $n_{\alpha}$ are monomials with the form $( \prod_{i=1}^d y_{1i}^{e_i} ) \cdot ( \prod_{i=1}^d y_{2i}^{f_i} )$ where $\sum_{i=1}^d e_i$ is odd for monomials $m_\alpha$ and even for monomials $n_\alpha$. From equation \ref{eq:antisymmterynewcoord}, we must have
    \begin{align*}
        -P_1(\by_1,\by_2)=P_1(-\by_1, \by_2) 
        \quad \mathrm{and} \quad
        -P_2(\by_1,\by_2) = P_2(-\by_1, \by_2), 
    \end{align*}
    and from the conditions on the monomials, $P_2(\by_1,\by_2)+ P_2(-\by_1, \by_2) = 2 P_2(\by) = 0$. This means $P_2 = 0$. Also, $P = P_1$, so each monomial $m_\alpha$ obeys equation \eqref{eq:antisymmterynewcoord}. Hence, $-m_\alpha(\by_1, \by_2) = m(-\by_1, \by_2)$ for any index $\alpha$, which is written explicitly as $- (\prod_{i=1}^d y_{1i}^{e_i})\cdot (\prod_{i=1}^d y_{2i}^{f_i}) = (-1)^{\sum_{i=1}^d e_i} \cdot (\prod_{i=1}^d y_{1i}^{e_i})\cdot (\prod_{i=1}^d y_{2i}^{f_i})$. This means $\sum_{i=1}^d e_i$ must be odd, so at least one of the $\{e_i\}_{i=1}^d$ must be odd. Let us now consider the case where $e_j$ is odd for some $j\in \{ 1, \ldots, d\}$. Here, we can factor out one $y_{1j}$ from $m_\alpha$, so $m_\alpha(\by) = y_{1j} q_\alpha(\by)$ where $q_\alpha(\by)=(\prod_{i=1, i \neq j}^d y_{1i}^{e_i})\cdot y_{1j}^{e_j-1} \cdot (\prod_{i=1}^d y_{2i}^{f_i})$. Because $\sum_{i=1}^d e_i -1$ must be an even number, we see that $q_\alpha$ has the following symmetry:
    \begin{align}
        q_\alpha(\by_1,\by_2) = q_\alpha(-\by_1, \by_2). 
    \end{align}
    In the original coordinate $\bx$, it simply means that $q_\alpha$ (expressed in $\bx$ coordinate) will be a totally symmetric monomial. Noting that $y_{1j}=x_{1j}-x_{2j} = D_j(\bx)$, we see $m_\alpha$, when expressed in $\bx$ coordinate, can be factored as $D_j(\bx)$ times a totally symmetric monomial. This factorization of one of $\{D_i(\bx)\}_{i=1}^d$ can be done for each monomial $m_\alpha(\bx)$, so this concludes our proof. 
\end{proof}

\subsubsection{Bounds for minimum module generators of \texorpdfstring{$n$}{n}-antisymmetric polynomials for \texorpdfstring{$n>2$}{ng2}}
\label{sec:bounds_on_generators_anti}

When we are representing $n$-antisymmetric polynomials, the use of Vandermonde determinants seem intuitive. We may mirror the previous Lemma \ref{lem:antisymetric_n_eq_2} for the general case $n > 2$ and $d \geq 2$; however, this will not hold true. One immediate reason is the degree of some $n$-antisymmetric functions. Recall the Vandermonde determinants $D_{l} (\bx) = \prod_{1 \leq i<j \leq n}(x_{i l}-x_{j l})$. Their degrees are $\binom{n}{2}$, so some $n$-antisymmetric functions of the same or lower degree cannot be represented this way. For example, when $n = 3$ and $d = 2$, one such function is $f(\bx_1, \bx_2, \bx_3) = (x_{22}-x_{12}) (x_{31}-x_{11})+(x_{12}-x_{32}) (x_{21}-x_{11})$ where $\deg(f) = 2 < 3 = \binom{3}{2}$.

This brings us to the realm of hardcore representation theory where finding the minimal number of generators for $n$-antisymmetric polynomials over the symmetric polynomial rings have been tackled by many eminent mathematicians over the years. For $n \geq 2$ and $d = 2$, the exact minimal number of such generators was found by a deep result of Haiman as a solution of the Haiman-Garsia conjecture. (See \citet{HaimanGarcia1996}, \cite{Haimansolo2003}, \citet{Haiman1994ConjecturesOT}, \citet{Bergalgcombcoinvspaces} for background. For explicit reference see \citet{haglund2004combinatorial}, page 2.) The solution would be the $n$-th Catalan number given by $\frac{(2n)!}{(n+1)!n!}$. For the record, when $n= d =2$ this solution $\frac{4!}{3!2!}=2$ agrees with our Lemma \ref{lem:antisymetric_n_eq_2}.    

Let $n \geq 2$ and $d \geq 1$, and let us follow the notations from Appendix \ref{sec:appendix2} from here onwards. The exact minimal number of $n$-antisymmetric generators is given by the dimension of the vector space $H_{\calS_n}(\R^n \otimes \R^d) \cap \calO(\R^n \otimes \R^d)_{\sgn}$, which is given by Lemma \ref{lemma:minimalgenasvectspacebasis}. As once stated, we aim to the determine the upper and lower bound of this minimal number of generators, and some great work has been done by \citet{Wallachpaperalternantsminimaldegree}; we will draw much from this work to find these bounds. For convenience, $n$-antisymmetric polynomials are now often called ``alternants,'' which is the same terminology used in \citet{Wallachpaperalternantsminimaldegree}.

Let us determine the lower bound first. Note that the the set of minimal degree alternants form a vector space over $\R$, and let this minimal degree be $r$. We will use this later on. In \citet[Theorem~1]{Wallachpaperalternantsminimaldegree}, the lower bound of the number of module generators for alternants over symmetric polynomials is given by the dimension of the vector space spanned by this set of minimal degree alternants. We provide an argument here, relating these two numbers: Suppose $\{f_1, f_2, \ldots, f_k\}$ is the minimal set of alternants that generate the $n$-antisymmetric polynomials as a module over the totally symmetric polynomials. Now, recall that any polynomial $g$ can be written as sum of homogeneous polynomials, i.e. $g = \sum_q g^{(q)}$, where $g^{(q)}$ is the homogeneous part with degree $q$. By Lemma \ref{lemma:homsub}, if $g$ is an alternant, then $g^{(q)}$ are also alternants. Now, suppose $\deg(g) = r$, which is the minimal degree in the set of minimal degree alternants, then $g$ must be homogeneous because of the minimality assumption. Consequently, $g = \sum_{j=1}^k u_j f_j = \sum_{j=1}^k (u_j f_j)^{(r)}$ for some totally symmetric polynomials $\{u_j\}_{j=1}^k$. Furthermore, for any $j \in [k]$, $(u_j f_j)^{(r)} = \sum_{a+b=r} u_j^{(a)}f_j^{(b)}$, where $a,b$ are nonnegative integers. In order to satisfy the minimality assumption for the degree $r$, every index $b = r$ where $f_j^{(b)} \neq 0$ and the corresponding $a = 0$. Hence, $g = \sum_{j = 1}^k u_j^{(0)} f_j^{(r)}$ where $u_j^{(0)} \in \R$ for $f_j^{(r)} \neq 0$, and so $g$ is represented by the dimension of the vector space spanned by degree $r$ alternants. (Also note that the minimal generating set of $n$-antisymmetric polynomials can be comprised of homogeneous polynomials, see Theorem \ref{thm:rep-antisymm-polys} in Appendix \ref{sec:appendix2}.)

Before stating the lower bound for the number of generators, recall a standard combinatorial fact: Given $d, n \in \N$, $n$ can be written uniquely as $n = \binom{r}{d} + j$ with $0 \leq j < \binom{r}{d-1}$ and $r \in \ZZ$. From \citet[Theorem~1]{Wallachpaperalternantsminimaldegree}, the minimum degree of the minimal generating set of alternants is $d \binom{r}{d+1} + j (r - d + 1)$; in the proof, the dimension of the vector space spanned by set of minimal degree alternants, which give the lower bound, is

\[
    \binom{\binom{r}{d-1}}{j} \ \text{where} \ n = \binom{r}{d} +  j \ \text{with} \ 0 \leq j < \binom{r}{d-1}.
\]

Now, we want to establish the upper bound for the minimum number of generators for the alternants over the totally symmetric polynomials. According to \citet{Wallachpaperalternantsminimaldegree}, the maximum degree of the minimial generating set of alternants is $\binom{n}{2}$, and this fact is proven in Lemma \ref{lemma:finitedimHGVgen}, Appendix \ref{sec:appendix2}. Now, mirroring \citet{Wallachpaperalternantsminimaldegree}, define a graded lexicographic order on the elements of $\ZZ^d$: For any $\bb, \bc \in \ZZ^d$, $\bb < \bc$ if $\sum_{i=1}^d b_i < \sum_{i=1}^d c_i$ or if $\sum_{i=1}^d b_i = \sum_{i=1}^d c_i$, then if $j$ being the first index where $b_j \neq c_j$, then $b_j < c_j$. For convenience, let us define $A \in \R^{d \times n}$ where $(A)_{ij} \in \ZZ$, and let $\bx = (\bx_1, \ldots, \bx_n) \in (\R^d)^n$ be the variables. Using a shorthand notation, we define a monomial as
\[
    \bx^A \coloneqq \prod_{i,j}^{d,n} x_{ij}^{a_{ij}}.
\]
Now, define $\alt(\cdot)$ on any function with the variables $\bx \in (\R^d)^n$ by
\[
    \alt(f)(\bx) \coloneqq \frac{1}{|\calS_n|} \sum_{\sigma \in \calS_n} \sgn(\sigma) f(\sigma.\bx),
\]
where, $\sigma.\bx = (\bx_{\sigma(1)}, \ldots, \bx_{\sigma(n)})$. Then, a basis for the alternants as a vector space over $\R$ is the set 
\[
    \Lambda = \left\{\alt(\bx^A): A = (\ba_1, \ldots, \ba_n) \in (\ZZ^d)^n \text{ and } \ba_1 < \ba_2 < \ldots < \ba_n \right\} 
\]
according to \citet[Proposition~4]{Wallachpaperalternantsminimaldegree}; of course, this is not the set of interest in our study. However, scalars in $\R$ are also totally symmetric polynomials, so $\Lambda$ with the additional condition of maximal degree of $\binom{n}{2}$, i.e. $\sum_{i=1,j=1}^{d,n} (A)_{ij} \leq \binom{n}{2}$, becomes a set of generators for the alternants over the totally symmetric polynomials, see Theorem \ref{thm:rep-antisymm-polys}. Let us call this set $\Lambda_{n,d}$, and so its cardinality is actually a loose upper bound on the minimum number of generators. Actually, finding this cardinality is a very difficult combinatorial problem, which might not have a closed-form expression; this is mainly due to this ordering $\ba_1 < \ba_2 < \ldots < \ba_n$ where $\{\ba_i\}_{i=1}^n$ are vectors in $\R^d$. Hence, for simplicity, we compute the number of non-negative integral solutions of $\sum_{i=1,j=1}^{d,n}(A)_{ij} \leq \binom{n}{2}$, which is the cardinality of a superset of $\Lambda_{n,d}$ without the ordering condition. The number of non-negative integral solutions give an upper bound
\[
    \left| \Lambda_{n,d} \right| \leq \binom{\binom{n}{2} + dn}{dn}
\]
on the minimum number of generators of $n$-antisymmetric polynomials as a module over the totally symmetric polynomials.

We tested our bounds in Table \ref{tab:bounds_and_exact_d2}, which shows the exact numbers of generators for $n \geq 2$ and $d=2$ falling (quite crudely) between the lower and upper bound. We also summarize our findings of Section \ref{sec:repofsymantisympol} in Table \ref{tab:summary_of_results}. We emphasize here again that the exact number or the bounds on the generators (of algebra or module) that we obtain here are all independent of any function that we are approximating and the degree of approximation as we are simply dealing with polynomials now.

\begin{table}[ht]
\centering
\begin{tabular}{ccccccccccc}
\multicolumn{11}{c}{Bounds and Exact Minimal Number of Generators for $d = 2$} \\ \hline
\multirow{3}{*}{$n$} &  & \multirow{3}{*}{$r$} &  & \multirow{3}{*}{$j$} &  & \multirow{2}{*}{Exact}                 &  & \multicolumn{3}{c}{Bounds}                                                    \\ \cline{9-11} 
                     &  &                      &  &                      &  &                                        &  & Lower                        &  & Upper                                       \\
                     &  &                      &  &                      &  & $\displaystyle \frac{(2n)!}{(n+1)!n!}$ &  & $\displaystyle \binom{r}{j}$ &  & $\displaystyle \binom{\binom{n}{2}+2n}{2n}$ \\ \hline
3                    &  & 3                    &  & 0                    &  & 5                                      &  & 1                            &  & 84                                          \\
4                    &  & 3                    &  & 1                    &  & 14                                     &  & 3                            &  & 3003                                        \\
5                    &  & 3                    &  & 2                    &  & 42                                     &  & 3                            &  & \num{1.85e5}                                \\
6                    &  & 4                    &  & 0                    &  & 132                                    &  & 1                            &  & \num{1.74e7}                                \\
7                    &  & 4                    &  & 1                    &  & 429                                    &  & 4                            &  & \num{2.32e9}                                \\
8                    &  & 4                    &  & 2                    &  & 1430                                   &  & 6                            &  & \num{4.17e11}                               \\ \hline
\end{tabular}
\caption{Demonstrate that the exact minimal number of generators fall between the derived bounds for $d = 2$. Note that $n = \binom{r}{2} + j$ where $j \in \{0, 1, \ldots, r-1\}$.}
\label{tab:bounds_and_exact_d2}
\end{table}

\begin{table}[ht]
\centering
\begin{tabular}{ccccccccccc}
\multicolumn{11}{c}{Summary of Results} \\ \hline
\multirow{2}{*}{Polynomial  Type}                 &  & \multirow{2}{*}{$n^*$} &  & \multirow{2}{*}{$d$} &  & \multirow{2}{*}{Exact}                 &  & \multicolumn{3}{c}{Bounds}                                                                                                   \\ \cline{9-11} 
                                                &  &                        &  &                      &  &                                        &  & Lower                                                      &  & Upper                                                        \\ \hline
$\text{Totally Symmetric}^\dag$                 &  & $\geq 2$               &  & $\geq 1$             &  & $\displaystyle \binom{n+d}{d}$         &  & --                                                         &  & --                                                           \\ \cline{1-1} \cline{3-3} \cline{5-5} \cline{7-7} \cline{9-9} \cline{11-11} 
\multirow{4}{*}{$n\text{-Antisymmetric}^\ddag$} &  & $\geq 2$               &  & $1$                  &  & $1$                                    &  & --                                                         &  & --                                                           \\
                                                &  & $2$                    &  & $\geq 1$        &  & $d$                                    &  & \multirow{3}{*}[-1.5mm]{$\displaystyle \binom{\binom{r}{d-1}}{j}$}&  & \multirow{3}{*}[-1.5mm]{$\displaystyle \binom{\binom{n}{2}+dn}{dn}$} \\
                                                &  & $>2$                   &  & $2$                  &  & $\displaystyle \frac{(2n)!}{(n+1)!n!}$ &  &                                                            &  &                                                              \\
                                                &  & $\geq 2$               &  & $\geq 2$             &  & --                                     &  &                                                            &  &                                                              \\ \hline
\end{tabular}
\caption{
$^\dag$ Totally symmetric polynomials forms an $\R$-algebra, so the number of algebra generators is given. $^\ddag$ $n$-Antisymmtric polynomials form a finitely generated module over the totally symmetric polynomials, so lower and upper bounds are given for the minimum number of module generators.. $^*$ In the antisymmetric case, $n = \binom{r}{d} + j$ for $0 \leq j < \binom{r}{d-1}$.}
\label{tab:summary_of_results}
\end{table}

\section{Acknowledgments and disclosure of funding}
S.G. and K.T. are grateful to their peers N. Ramachandran, H. Bhatia, and S. Bhattacharya, in the Dept. of Mathematics, UCSD for enlightening conversations about many algebraic facts. S.G. is thankful to S. Chhabra for introducing him to the broad area of mathematical deep learning. The authors are grateful to Prof. Steven Sam and Prof. Brendon Rhoades for illuminating conversations on commutative algebra and representation theory. R.S. would like to thank the Institute for Pure and Applied Mathematics, Stanford for being generous hosts, and for providing a great environment during the period when most of this work was completed. The authors would also like to thank Prof. Nolan Wallach, for communicating many of the results and proofs related to invariant and representation theory, which have been used extensively in Appendix~\ref{sec:appendix2}. The authors declare that they have no competing interests.

\appendix 
\section{Necessary background for commutative algebra}
\label{sec:appendix}

In this section, some basic definitions and theorems from commutative algebra are given, which are useful in understanding the results in Section \ref{sec:repCkantisympol} and Appendix~\ref{sec:appendix2}. For more details, the reader is referred to any standard textbook such as \citet{AtiyahMacDcommalgebrabook}, and the results will be referenced. We assume the reader has a basic understanding of groups, rings, and vector spaces over fields. Below the operations of ring addition and multiplication are denoted using the symbols $+$ and $\cdot$ respectively. We will also use these same symbols later to denote the group addition and scalar multiplication operations for a module (defined below in Definition~\ref{def:RModule}), but the distinction will always be clear from context, and no confusion should arise.

If $R$ is a ring, for any $r,s \in R$, we will frequently write $rs$ to mean $r \cdot s$, when there is no chance for confusion. A ring is \textit{unital} if multiplication has an identity element (denoted as $1$). A ring is called \textit{commutative} if $r \cdot s = s \cdot r$ for all $r,s \in R$. Throughout this paper, we will assume our rings to be unital, commutative rings. A nonzero commutative ring in which every nonzero element has a multiplicative inverse is called a field, e.g. $\R$ and $\mathbb{C}$. A nonzero commutative ring $R$ is called an \textit{integral domain} if $x \cdot y =0$, for $x, y \in R$, implies either $x$ or $y$ is the zero element of $R$. For example, if $\{\bx_i\}_{i=1}^{n} \subset \R^d$, then the set of polynomials $\mathbb{R}[\bx_1, \ldots, \bx_n]$, form a ring when equipped with the usual addition and multiplication operations for polynomials. This polynomial ring plays a key role in this paper. In fact, $\mathbb{R}[\bx_1, \ldots, \bx_n]$ is also an integral domain \citep[Page~2]{AtiyahMacDcommalgebrabook}. Any subring of $\mathbb{R}[\bx_1, \ldots, \bx_n]$ inherits the same addition and multiplication operations from $\mathbb{R}[\bx_1, \ldots, \bx_n]$. A notion that we will repeatedly encounter is that of an \textit{ideal} of a ring:

\begin{definition}[Ideal of a ring]
\label{def:ideal}
    An ideal $I$ of a commutative ring $(R,+,\cdot)$ is a subset of $R$ such that $(I,+)$ is a subgroup of $(R,+)$, and for every $r \in R$ and $x \in I$, the product $r \cdot x \in I$. 
\end{definition}

Another important class of objects that we will also encounter frequently is the notion of a \textit{module} over a ring, which generalizes the concept of a vector space over a field.

\begin{definition}[Module over a ring]
\label{def:RModule}
    Given a ring $R$, a set $M$ is called an $R$-Module if $(M,+)$ is an abelian group, equipped with an operation $\cdot: R \times M \to M$ satisfying $r \cdot (x + y) = r \cdot x + r \cdot y$, $(r+s)\cdot x = r \cdot x + s \cdot x$, $(rs) \cdot x = r \cdot (s \cdot x)$, $1 \cdot x = x$, for all $r,s \in R$ and $x, y \in M$. The operation $\cdot$ is called scalar multiplication.
\end{definition}

For a $R$-module $M$, the group identity will also be denoted as $1$, and again the distinction from the ring unit will be clear from context. If $M$ is an $R$-module and $N$ is a subgroup of $M$, then $N$ is a $R$-submodule if for any $n \in N$ and any $r \in R$, $r \cdot n \in N$. Recall that if $I$ is an ideal of a ring $R$, then $I$ is naturally an $R$-module. An $R$-module $M$ is \textit{finitely generated} if there exist finitely many elements $\{ y_i\}_{i = 1}^n \subset M$ such that every element of $M$ is a linear combination of $\{ y_i\}_{i = 1}^n $ with coefficients from $R$.


We next discuss \textit{Noetherian rings} and \textit{Noetherian modules} which play a key role in Section~\ref{sec:repCkantisympol}, but we will not give the axiomatic definition of these objects (which depends on ascending chain conditions). Alternate and equivalent definitions are given below because they are more relevant to the material in this paper.

\begin{definition}[Noetherian ring]
\label{def:noetherian_ring}
    A ring $R$ is called Noetherian if every ideal of $R$ is finitely generated as a $R$-module.
\end{definition}

\begin{definition}[Noetherian module]
\label{def:noetherian_module}
    A $R$-module $M$ is called Noetherian if every submodule of $M$ is finitely generated over $R$.
\end{definition}

For the characterization of Noetherian modules that we gave above in Definition~\ref{def:noetherian_module},  the reader is referred to \citet[Proposition~6.2]{AtiyahMacDcommalgebrabook}. Finally, we define the notion of an \textit{associative algebra} over a ring, which is also extensively used in this paper.  

\begin{definition}[Associative algebra over a ring]
\label{def:algebra}
    Let $R$ be a ring, and let $\calA$ be a $R$-module. Then $\calA$ is called an $R$-algebra (or an algebra over $R$), if $\calA$ also forms a ring such that the ring addition is the same operation as module addition, and module scalar multiplication $\cdot$ satisfies $r \cdot (x \ast y) = (r \cdot x) \ast y = x \ast (r \cdot y)$ for all $r \in R$ and $x,y \in \calA$, where $\ast$ denotes ring multiplication in $\calA$. 
\end{definition}
When the associative algebra ring multiplication $\ast$ is also commutative, we say that it is a \textit{commutative algebra}. All the associative algebras that we will encounter in this paper are commutative algebras, and hence for simplicity we will simply refer to them as \textit{algebras} going forward. An $R$-algebra $\calA$ is \textit{finitely generated} if there exists a finite set of elements $\{z_i\}_{i=1}^m \subset \calA$ such that any element of $\calA$ can be written as a finite linear combination of terms, with coefficients in $R$, where each term is a finite product of the elements in $\{z_i\}_{i=1}^m$. For example, the polynomial ring $\mathbb{R}[\bx_1, \ldots, \bx_n]$ introduced previously, forms an $\mathbb{R}$-algebra. It is in fact finitely generated by the set $\{ x_{ij} \}_{i=1,j=1}^{n,d}$. 




Another notion that is needed in Section~\ref{sec:repCkantisympol}, is that of \textit{integral} elements in a ring, over a subring. Next we define this notion, and then state an important result involving  integral elements (Proposition~\ref{prop:prp1appen} below, the proof of which can be found in \citet[Proposition~5.1]{AtiyahMacDcommalgebrabook}).

\begin{definition}[Integral element over subring]
\label{def:integral}
    Let $R$ be a ring and $S \subset R$ be a subring. An element $x \in R$ is integral over $S$, if $x$ is a root of a monic polynomial with coefficients in $S$. Here monic polynomial means that the polynomial is univariate and the coefficient of the highest degree term of the polynomial is $1$. We say that $R$ is integral over $S$ if every element $x \in R$ is integral over $S$.
\end{definition}

\begin{proposition}
\label{prop:prp1appen}
    Let $R$ be a ring and $S \subset R$ be a subring. Then $x \in R$ is integral over $S$ if and only if $S[x]$ is a finitely generated $S$-module, where $S[x]$ is the subring generated by $S \cup \{x\}$.
\end{proposition}

We also need the following three results, the proofs of which can be found in \citet[Corollary~7.6]{AtiyahMacDcommalgebrabook}, \citet[Theorem~3.7.i]{MatsumuraCommutativeringtheorybook}, and \citet[Proposition~6.5]{AtiyahMacDcommalgebrabook} respectively. Of these, Hilbert's basis theorem is a famous theorem in commutative algebra.

\begin{theorem}[Hilbert's basis theorem]
\label{thm:Hilbert's basis theorem}
    If $R$ is a Noetherian ring then $R[x_1, \ldots, x_n]$ is also a Noetherian (polynomial) ring. 
\end{theorem}

\begin{theorem}[Eakin-Nagata theorem]
\label{thm:eakinnagata}
    If $R$ is a Noetherian ring and $S$ is a subring such that $R$ is finitely generated as a $S$-module, then $S$ is also a Noetherian ring.  
\end{theorem}

\begin{proposition}
\label{prp2appen}
    If $R$ is a Noetherian ring, and $M$ is a finitely generated $R$-module, then $M$ is a Noetherian $R$-module. 
\end{proposition}

\section{Necessary results from representation theory}
\label{sec:appendix2}

In this section, we establish some facts in representation theory, which are used in Section~\ref{sec:repCkantisympol} to prove the upper and lower bounds of the minimum number of module generators for the set of $n$-antisymmetric polynomials (as a module) over the totally symmetric polynomials. In particular, these polynomials belong to the ring $\R[\bx_1, \ldots, \bx_n]$ where $\{\bx_i\}_{i=1}^n \subset \R^d$.
We are grateful to Prof. Nolan Wallach for communicating and discussing these results with us. The results are included here for completeness and may illuminate the readers on the main results of our paper, especially those from the machine learning community. Finally, the results are discussed here for the field $\R$; however, they remain true for polynomials over any field of characteristic zero. For a more detailed discussion of the topics presented here, the reader can refer to \citet[Section~3.7.6]{Wallachbook2017}. 

This section is structured as follows: In Section~\ref{subsec:basbackrepinv}, we introduce the $\R$-algebra of polynomials equipped with an inner-product, identify several modules and subalgebras associated to a group action on the space of polynomials, and state some of their properties. Section~\ref{subsec:finitedimhgv} is dedicated to proving the finite-dimensionality of the $\R$-vector space $H_{\calS_n}(\R^n \otimes \R^d)$, which is an important subspace of the vector space of polynomials $\mathbb{R}[\bx_1,\dots,\bx_n]$ that plays a key role in the analysis. Then in Section~\ref{subsec:modstrantisym}, we prove several structural results for this subspace, eventually culminating with Lemma~\ref{lemma:minimalgenasvectspacebasis}, which is the main tool for quantifying the minimum number of module generators for the space of $n$-antisymmetric polynomials. In these subsections, we first prove the results for a general group action $G$, then state the results for the special case $G=\calS_n$.
The precise definition of the group action $\calS_n$ relevant for us is provided below in Section~\ref{subsec:basbackrepinv}.

\subsection{Polynomial algebras and modules induced by a group action}
\label{subsec:basbackrepinv}

We begin by defining the notion of representation of a group:
\begin{definition}[Representation of a group]
    A representation of a group $G$ on a vector space $Y$ over a field $\mathbb{K}$ is a group homomorphism $\rho$ from $G$ to $GL(Y)$, the general linear group on $Y$. This means $\rho:  G \to GL(Y)$ satisfies $\rho(g_1g_2)=\rho(g_1)\rho(g_2)$ for all $g_1, g_2 \in G$. The dimension of $Y$ is called the dimension of the representation. 
\end{definition}

Here, we will only deal with finite-dimensional representations. From this definition, it is clear that representations can also be interpreted as group actions. Now since elements of $GL(Y)$ can be represented by invertible matrices, we can take their trace, and the resulting quantity defines the \textit{character} of the representation, denoted as $\chi$. Thus $\chi(g) \coloneqq \tr(\rho(g))$ for all $g \in G$, if $\rho$ is a representation of  a group $G$. A subspace $W$ of $Y$ is called $G$-invariant if $\rho(g)w \in W$ for all $g\in G$ and $w \in W$. A $G$-representation is called \textit{irreducible} if $\dimension(Y) \ne 0$ and the only $G$-invariant subspaces of $Y$ are $\{0\}$ and $Y$ itself. Interested readers are referred to \citet[Chapter~1]{Brucesagansymmetricgroup} for more information. 

Let $V$ be any finite dimensional inner product space over $\R$ where $\dimension(V) = m$. Now, we will define a few polynomial $\mathbb{R}$-algebras and $\mathbb{R}$-modules over $V$, equipped with an inner-product, that we will use extensively. Let $ \{ v_i \}_{i = 1}^m$ be an orthonormal basis of $V$. This allows us to define coordinates in $V$ by expressing any vector $v \in V$ as $v= \sum_{i=1}^m z_i v_i$, where $z_i \in \R$ for all $i \in [m] = \{1, 2, \ldots, m\}$. Now, any $v \in V$ can be denoted by its coordinates $\bz=(z_1, \ldots, z_m)$. We define an $\mathbb{R}$-algebra of polynomials on $V$ with respect to these coordinates, which we will denote as $\calO(V)$, i.e. $\calO(V) := \mathbb{R}[\bz]$. We also define a positive definite inner product on $\calO(V)$ as 
\begin{align*}
    \langle f, g \rangle_{\calO(V)} = f\left(\frac{\partial}{\partial z_1}, \ldots, \frac{\partial}{\partial z_m} \right)g(z_1, \ldots, z_m) \bigg \lvert_{\bz = 0}, \;\; \forall f,g \in \calO(V),
\end{align*}
which turns $\calO(V)$ into an inner-product space. We also introduce a useful notation: If $W,W'$ are subsets of $\calO(V)$ then we define $WW'$ as the set of all finite sums of the form $\sum_{i=1} a_i b_i$, where each $a_i \in W$ and $b_i \in W'$. Obviously, $WW'=W'W$. We note a special case where $W$ is a subring of $\calO(V)$, and $W'$ is a subset of $\calO(V)$, then $WW'$ is the $W$-module generated by $W'$ and is a $W$-submodule of $\calO(V)$. 

Next, let $G$ be a finite subgroup of $GL(V)$ acting on $\calO(V)$ as $s . f(\bz) \coloneqq s . f(z_1, \ldots, z_m)= f(s^{-1} \bz)$ 
for any $f \in \calO(V)$, $\bz \in V$, and $s \in G$. Let us denote $\calO(V)^G$ as the $\mathbb{R}$-algebra of polynomials invariant under $G$ action, i.e. $\{f \in \calO(V): s. f(z)=f(z),\; \forall \ s \in G\}$, and it is easy to check that it is indeed an $\mathbb{R}$-algebra. We will call elements of $\calO(V)^G$ as \textit{invariants} in this section -- where the group $G$ will be specified and understood from context -- to avoid conflict with the definition of $G$-invariant functions in Definition~\ref{def:G-invariantfn}. Denote $\calO_+(V)^G$ as the subspace of invariants that vanish at $\bz = 0$. One can then define $H_G(V)$ to be the orthogonal complement of $\calO(V)\calO_+(V)^G$ in $\calO(V)$, where $\calO(V)\calO_+(V)^G$ can be easily shown to be the smallest ideal of $\calO(V)$ containing $\calO_+(V)^G$. The space $H_G(V)$ is called the space of \textit{$G$-harmonic} polynomials, and it is a known result that it is equal to the set of all polynomials annihilated by all $G$-invariant constant coefficient differential operators on $V$ \citep[Lemma~3.105]{Wallachbook2017}. Finally, if $\chi$ is a character of an irreducible representation of $G$, then we can define $\calO(V)_{\chi} \coloneqq \{f \in \calO(V) : f(g.v)=\chi(g) f(v), \;\; \forall v \in V, g \in G\}$.

Now let us bring the context of our work into these definitions and notations. For Section ~\ref{sec:repCkantisympol}, we can take $V=\R^n \otimes \R^d$ and $G = \calS_n$, which acts on the first factor of the tensor product. We identify $\R^n \otimes \R^d$ with $(\R^d)^n$, and recall the notation of an element in $(\R^d)^n$ -- denoted as $\bx = (\bx_1,\dots,\bx_n)$ for each $\bx_i \in \R^d$. We maintain the same ordering of coordinates under the identification $\R^n \otimes \R^d \cong (\R^d)^n$, so that $\bx$ gives coordinates on $\R^n \otimes \R^d$. In this case, $\calO(V)^{\calS_n}$ is precisely the set of totally symmetric polynomials, and $H_{\calS_n}(V)$ is the set of polynomials in $\{x_{ij} : i \in [n], j \in [d]\}$ that are annihilated by all operators of the form $\sum_{i=1}^n \partial_{x_{i1}}^{a_1} \cdots \partial_{x_{in}}^{a_n}$ where $\partial_{x}^a = \frac{\partial^a}{\partial x^a}$ and $0 \neq (a_1,\dots,a_d) \in \mathbb{N}^d$. Similarly, if we take $\chi$ to be the character of the sign representation of $\calS_n$, the set of polynomials $\calO(V)_{\chi}$, denoted as $\calO(V)_{\sgn}$, are the $n$-antisymmetric polynomials that we defined previously in Section~\ref{sec:notation}. (The sign representation of $\calS_n$ is the one-dimensional representation of $\calS_n$ defined by $\rho(\sigma) = \sgn(\sigma)$, where $\sgn(\sigma)$ is the signature of the permutation $\sigma \in \calS_n$.) We note that $\calO(V)_{\sgn}$ forms an $\calO(V)^{\calS_n}$-module.

We now prove an important property of $\calO(V)_{\sgn}$ in the next lemma. One should note that a similar result holds in the general case of arbitrary $V$, $G$, and $\chi$, and the proof changes slightly.

\begin{lemma}
\label{lem:finitegenatisymonsym}
    For $V=\R^n \otimes \R^d$, $G = \calS_n$, and $\calO(V)=\mathbb{R}[\bx_1, \ldots, \bx_n]$, $\calO(V)_{\sgn}$ forms a finitely generated module over $\calO(V)^{\calS_n}$. 
\end{lemma}
\begin{proof}
Recalling Definition~\ref{def:integral}, we can show that $\calO(V)$ is integral over $\calO(V)^{\calS_n}$ (for a proof, see \citet[Chapter~5, Exercise~12]{AtiyahMacDcommalgebrabook}). In fact, we observe that for any $f \in \calO(V)$, $f$ is a root of the following univariate polynomial in variable $t$: $\prod_{\sigma \in G}(t - \sigma.f)$. This polynomial is monic in $t$ and the coefficients of $t$ in this polynomial all belong to $\calO(V)^{\calS_n}$. Now clearly $\calO(V)$ is a finitely generated $\mathbb{R}$-algebra, with a generating set given by $\{ x_{ij} : i \in [n], \; j \in [d] \}$. We just showed above that each $x_{ij}$ is integral over $\calO(V)^{\calS_n}$. Then letting $S=\calO(V)^{\calS_n}$ and $R=\calO(V)$ in Proposition~\ref{prop:prp1appen}, we get that the subring of $\calO(V)$ generated by $\calO(V)^{\calS_n}$ and $\{ x_{ij} : i \in [n], \; j \in [d] \}$ is a finitely generated $\calO(V)^{\calS_n}$-module. However, the subring generated by $\calO(V)^{\calS_n} \cup \{ x_{ij} : i \in [n], \; j \in [d]\}$ is $\calO(V)$, and so we have proved that $\calO(V)$ is a finitely generated $\calO(V)^{\calS_n}$-module.    

Next, we note that $\calO(V)$ is a commutative, Noetherian ring, and this is a direct consequence of Theorem~\ref{thm:Hilbert's basis theorem}. Since $\calO(V)^{\calS_n} \subset \calO(V)$ is a commutative ring (by virtue of being an $\R$-algebra), and $\calO(V)$ is a finitely generated $\calO(V)^{\calS_n}$-module, it then follows that $\calO(V)^{\calS_n}$ is a Noetherian ring by Theorem \ref{thm:eakinnagata}. Finally, by Proposition~\ref{prp2appen}, we can conclude that $\calO(V)$ is a Noetherian module over $\calO(V)^{\calS_n}$, and thus every $\calO(V)^{\calS_n}$-submodule of $\calO(V)$ is finitely generated. Since the $n$-antisymmetric polynomials $\calO(V)_{\sgn}$, form an $\calO(V)^{\calS_n}$-submodule of $\calO(V)$, the proof is complete.
\end{proof}

\vspace*{-0.5cm}
\subsection{Finite-dimensionality of \texorpdfstring{$H_{\calS_n}(\R^n \otimes \R^d)$}{}}
\label{subsec:finitedimhgv}

Our next goal is to prove that $H_{\calS_n}(\R^n \otimes \R^d)$ introduced in the previous subsection is a finite-dimensional $\R$-vector space. To proceed, we need to introduce some more terminology, and let us work more generally for arbitrary $V$ and finite group $G$ of $GL(V)$. As before in Section~\ref{sec:bounds_on_generators_anti}, let $f^{(j)}$ denote the homogeneous component of $f \in \calO(V)$ of degree $j$. Then for any polynomial $f \in \calO(V)$ of degree $k$, we can write $f =\sum_{j=0}^k f^{(j)}$. If $W$ is a subspace of $\calO(V)$ such that $f \in W$ implies $f^{(j)} \in W$ for all $j$, then we say that $W$ is a \textit{homogeneous subspace}. Given a homogeneous subspace $W$ of $\calO(V)$, we can write $W=\bigoplus_{j}W^{(j)}$, with $W^{(j)} := \{f^{(j)}: f \in W\} \cup \{0\}$. Let us first prove some properties about homogenous subspaces, and identify a few that are important for us in the next lemma:

\begin{lemma}
\label{lemma:homsub}
Let $W$ be a subring of $\calO(V)$ that is also a homogenous subspace, and let $G$ be a finite subgroup of $GL(V)$ acting on $\calO(V)$. Then we have the following:
\begin{enumerate}[label=(\alph*)]
    \item If $W'$ is a homogenous subspace of $\calO(V)$, then $WW'$ is also a homogenous subspace.
    \item Suppose $A_1, A_2, A_3$ are subspaces of $\calO(V)$, and $A_1 = A_2 + A_3$. Then if any two of them are homogenous subspaces of $\calO(V)$, then so is the third.
    \item $\calO(V)^G$, $\calO(V)^G_+$, $\calO(V)_{\chi}$, $\calO(V)\calO(V)^G_+$, $H_G(V)$ are homogeneous subspaces of $\calO(V)$. 
\end{enumerate}
\end{lemma}

\begin{proof}
(a) Every element $f \in WW'$ is of the form $\sum_{i=1} a_i b_i$, where each $a_i \in W$ and $b_i \in W'$. Then for any $p \in \mathbb{N}$, we have
\begin{align*}
     f^{(p)}= \sum_{i=1}\sum_{\substack{j,k \geq 0 \\ j + k=p}} a_{i}^{(j)} b_{i}^{(k)}.
\end{align*}  
Since each term in the summation is in $WW'$, as $W,W'$ are homogeneous, we get that $f^{(p)} \in WW'$. 

(b) First assume that $A_2$ and $A_3$ are homogenous subspaces, and let $f \in A_1$. Then for any integer $j \geq 0$, $f_2 \in A_2$, and $f_3 \in A_3$, we have $f^{(j)} = f_2^{(j)} + f_3^{(j)}$. From this, we conclude that $f^{(j)} \in A_1$, as both $f_2^{(j)} \in A_2$ and $f_3^{(j)} \in A_3$, by homogeneity. The other cases follow by a similar argument.

(c) If two polynomials are equal, then so are each of their homogeneous components. For $s \in G$, if we write the group action as $s.f(\bz) = f(s^{-1}.\bz)$, then for $f=\sum_j f^{(j)}$, $(s.f)^{(j)}=s.(f^{(j)})$ for all $j$. This observation proves that $\calO(V)^G$, $\calO(V)^G_+$, $\calO(V)_{\chi}$ are homogeneous subspaces. Then by (a) we see that $\calO(V)\calO(V)^G_+$ is a homogeneous subspace since $\calO(V)$ is clearly homogenous. By (b), we conclude that $H_G(V)$ is a homogeneous subspace because $\calO(V) = \calO(V)\calO_+(V)^G \oplus H_G(V)$.
\end{proof}

We may now specialize to the context of our work, and assume that $V = \R^n \otimes \R^d$ and $G = \calS_n$. First let $\by = (y_1, \ldots, y_m) \in \R^m$ and $\ba = (a_1, \ldots, a_m) \in \ZZ^m$. We define 
\[
    h_{k,m}(\by) 
    \coloneqq \sum_{a_1 + \ldots + a_m = k} y_{1}^{a_1} y_{2}^{a_2} \cdots y_{m}^{a_m} 
    = \sum_{|\ba| = k} \by^{\ba}
\]
along with the shorthand notations $|\ba| = a_1 + \ldots + a_m$ and $\by^{\ba} = y_{1}^{a_1} y_{2}^{a_2} \cdots y_{m}^{a_m}$. We also let $>_{\text{lex}}$ be the lexicographic order on the monomials $\{y_j\}_{j=1}^m$ such that $y_1 > y_2 > \ldots > y_m$. Then the ideal, denoted as $\calI \subset \R[y_1, \ldots, y_m]$, of symmetric polynomials of positive degree is generated by $\eta_{k,m}(\by) = \sum_{i=1}^m y_i^k$ for $k \in [m]$. A Gr\"{o}bner basis for $\calI$ is given by the polynomials
\begin{equation}
    \{ u_{k,m}(\by) = h_{k,m-k+1}(y_k, \ldots, y_n)\}_{k=1}^m.
\end{equation}
More specifically, these polynomials are in the ideal $\calI$. Readers can refer to \citet[Proposition~2.1]{mora2003grobner} for this fact. 

Now $V=\R^n \otimes \R^d$ can be identified as a vector space of real valued $n \times d$ matrices $M_{n \times d}(\R)$. Let $\calS_n$ be identified as $n \times n$ permutation matrices, which acts from the left by multiplication. Consequently, we need to define a lexicographic order $>_{\text{lex}}$ on $\{x_{ij} : i \in [n], j \in [d]\}$: $x_{11} > x_{12} > \ldots > x_{1d} > \ldots > x_{n1} > x_{n2} > \ldots > x_{nd}$. We note this lexicographic ordering of the variables also defines a total ordering of the monomials of degree $k$ in these variables, for every integer $k$. Let us define the family of polynomials $y_{\ell,\bz}(\bx) := \sum_{j=1}^d z_j x_{\ell j}$ for every $\bz = (z_1, \ldots, z_d) \in \R^d$ and $\ell \in [n]$. Consequently, given $\sigma \in \calS_n$, its action is defined as $\sigma . y_{\ell,\bz}(\bx) \coloneqq y_{\sigma^{-1}(\ell),\bz}(\bx)$. We make the following claim:

\begin{claim}
\label{claim:vkninideal}
    Let $V=\R^n \otimes \R^d$ and $G = \calS_n$. Then for every $\bz \in \R^d$, the ideal $\calI_{n,d} \coloneqq \calO(V) \calO_+(V)^{\calS_n}$  contains the set of polynomials 
    \begin{equation}
    \label{eq:v-knz-def}
    \left\{ v_{k,n,\bz} := h_{k,n-k+1} (y_{k,\bz}, y_{k+1,\bz}, \ldots, y_{n,\bz}) \right\}_{k=1}^n.
    \end{equation}
\end{claim}

\begin{proof}
    Let us fix an arbitrary $\bz \in \R^d$ and note that the ring of real polynomials in $\{y_{1,\bz},\dots,y_{n, \bz}\}$ is a subring of $\calO(V)$. Also, every non constant $\calS_n$-symmetric polynomial in $\{y_{1,\bz},\dots,y_{n, \bz}\}$ is particularly a non-constant, totally symmetric polynomial in $\calO(V)$, i.e. it is in $\calO_+(V)^{\calS_n}$. This makes the ideal generated by the symmetric polynomials of positive degree in the ring $\R[y_{1, \bz},\dots,y_{n, \bz}]$ a subset of $\calI_{n,d}$. Then the observations from \citet{mora2003grobner}, mentioned above, completes the proof.
\end{proof}

The set of polynomials defined in~\eqref{eq:v-knz-def} can be used to define another important family of polynomials $p_{k,n,\bb} \in \R[\bx_1,\dots,\bx_n]$, where $\bb = (b_1,\dots,b_d)$ is a $d$-tuple of non-negative integers satisfying $\sum_{\ell=1}^d b_{\ell}=k$. They are obtained by the expansion of $v_{k,n,\bz}$ (using the definition of $h_{k,n-k+1}$):
\begin{equation}
\label{eq:p-knb-def}
\begin{split}
    v_{k,n,\bz}(\bx_k, \ldots, \bx_n)
    &= \sum_{a_k + \ldots + a_n = k} \left( \sum_{j=1}^d z_j x_{kj} \right)^{a_k} \left( \sum_{j=1}^d z_j x_{(k+1)j} \right)^{a_{k+1}} \cdots \left( \sum_{j=1}^d z_j x_{nj} \right)^{a_{n}} \\
    &= \sum_{b_1 + \ldots + b_d = k} z_1^{b_1} z_2^{b_2} \cdots z_d^{b_d} p_{k,n,\bb}(\bx_k, \ldots, \bx_n).
\end{split}
\end{equation}
In particular, $p_{k,n,\bb}$ are homogeneous polynomials of degree $k$; we also infer that each $p_{k,n,\bb}$ is in the ideal $\calI_{n,d}$, which will be proven using Lemma~\ref{lem:det-nonzero}. First, for a fixed $k \in [n]$, the set $\{ \bb=(b_1, b_2, \ldots, b_{d}) \in \ZZ^d : |\bb| = k\}$ has cardinality $N = \binom{k+d-1}{d-1}$, which is the  number of polynomials $p_{k,n,\bb}$ in the expression $v_{k,n,\bz}$. Let us enumerate them as $\bb_1, \bb_2, \ldots, \bb_N$. Since we can choose $\bz$ freely from $\R^d$, we also have the following lemma:

\begin{lemma}
\label{lem:det-nonzero}
    There exist points $\bz_1, \bz_2, \ldots, \bz_N \in \R^d$ such that the $N \times N$ matrix $A_N$, whose entries are given by $(A_N)_{ij} := \bz_i^{\bb_j}$, is invertible.
\end{lemma}

\begin{proof}
Let us fixed a $\bz = (p_1,\dots,p_d) \in \R^d$, which will be determined later. Define $\bz_i = (p_1^{i-1}, \ldots, p_d^{i-1})$ and $q_i = \bz^{\bb_i}$ for all $i \in [N]$.
From these choices, $A_N^{\top}$ is a Vandermonde matrix, and so $(\det A_N)(\bz) = \prod_{1 \leq i < j \leq N}(q_j-q_i)$. This vanishes if and only if $q_i=q_j$ for some $i \ne j$, and since we want $A_N$ to be invertible, we need to choose $\bz$ such that $\prod_{1 \leq i < j \leq N}(q_j-q_i) \ne 0$. It is sufficient to choose $\bz$ such that $\prod_{1 \leq i < j \leq N}(q_j-q_i)$ is not identically the zero polynomial in $\R[\bz]$. 

The polynomial $(\det A_N)(\bz)$ can be identically zero if and only if at least one of its factors is identically zero polynomial, i.e. $q_j = q_i$ (as polynomials)  for some $i \ne j$. (Recall that $\R[\bz]$ is an integral domain (see appendix \ref{sec:appendix}). This means that $fg = 0$ for some $f,g \in \R[\bz]$ if and only if $f=0$ or $g=0$.) However, given that every $\bb_i$ in the enumeration are distinct from each other, we must have $q_i=\bz^{\bb_i} \neq \bz^{\bb_j}=q_j$ (as polynomials), whenever $i \neq j$. It means that there is at least one choice of $\bz \in \R^d$ such that $(\det A_N) (\bz) \neq 0$. 
\end{proof}
Let us choose and fix values of $\bz_1, \bz_2, \ldots, \bz_N$ to get an invertible matrix $A_N$ as in Lemma~\ref{lem:det-nonzero}. Then for fixed $k$ and $n$ using \eqref{eq:p-knb-def} we obtain a system of equations, written as $Y = A_N X$, where $Y$ is the vector formed by the polynomials $\{ v_{k, n, \bz_i} \}_{i=1}^{N}$, and $X$ is the vector formed by the polynomials $\{p_{k,n,\bb_i} \}_{i=1}^{N}$. Since every entry of $Y$ is in the ideal $\calI_{n,d}$ (by Claim \ref{claim:vkninideal}), then every entry of the vector $X = A_N^{-1}Y$ is also in the ideal $\calI_{n,d}$. This finishes the proof that each $p_{k,n,\bb} \in \calI_{n,d}$.

Next, notice that the leading monomial of $p_{k,n,\bb}$ based on our lexicographic ordering is $\bx_k^{\bb} = x_{k1}^{b_1} x_{k2}^{b_2} \cdots x_{kd}^{b_d}$, with a coefficient
\[
    C_{k,\bb} = \frac{k!}{b_1! \cdots b_d!}.
\]
This is true because the leading monomial requires $a_k = k$, which implies that $a_j = 0$ for $k < j \leq n$. Finally, we can now prove our intended result:

\begin{lemma}
\label{lemma:finitedimHGVgen}
Let $V = \R^n \otimes \R^d$. If $f$ is a homogeneous polynomial in $\calO(V)$ of degree $r > \binom{n}{2}$, then $f \in \calO(V)\calO_+(V)^{\calS_n}$.
\end{lemma}

\begin{proof}
 We prove this by contradiction. Assume that there is at least one homogeneous polynomial of degree $r > \binom{n}{2}$ 
 which is not in $\calI_{n,d} = \calO(V) \calO_+(V)^{\calS_n}$. Among all such homogenous polynomials of degree $r$, choose $f$ such that its leading monomial is minimal with respect to our lexicographic ordering. Suppose the leading monomial of $f$ is
    \[
        c \cdot \bx_1^{\ba_1} \bx_2^{\ba_2} \cdots \bx_n^{\ba_n},
    \]
    where $|\ba_1| + \ldots +|\ba_n| = r = \text{deg}(f)$, and $c \in \R$. We note that there must exist a first $k \in [n]$ such that $|\ba_k| \geq k$ (otherwise, if $|a_j| < j$ for all $j \in [n]$, then $r = |\ba_1| + \ldots +|\ba_n| \leq \sum_{j=1}^n (j-1) = \binom{n}{2}$, contradicting our initial assumption on $\text{deg}(f)$). We fix this $k \in [n]$ and write $\bx_k^{\bm{a}_k} = \bx_k^{\bm{\alpha}} \bx_k^{\bm{\beta}}$, for some $\bm{\alpha}, \bm{\beta} \in \ZZ^d$ with $|\bm{\beta}|=k$. Finally, let us consider the following polynomial: 
    \[
        g = \frac{c \cdot \bx_{1}^{\ba_1} \bx_{2}^{\ba_2} \cdots \bx_{n}^{\ba_n}}{C_{k,\bm{\beta}} \cdot \bx_k^{\bm{\alpha}} \bx_k^{\bm{\beta}} } \bx_k^{\bm{\alpha}} p_{k,n,\bm{\beta}},
    \]
    where the leading monomial of $\bx_k^{\bm{\alpha}} p_{k,n,\bm{\beta}}$ is $ C_{k,\bm{\beta}} \cdot \bx_k^{\bm{\alpha}} \bx_k^{\bm{\beta}}$. 
    Thus the leading monomial of $g$ is the leading monomial of $f$. Since $p_{k,n,\bm{\beta}} \in \calI_{n,d}$, we have $g \in \calI_{n,d}$ implying $0 \neq f - g \notin \calI_{n,d}$. However, according to this construction, the leading monomial of $f - g$ is less than the leading monomial of $f$ by the lexicographic order. This contradicts the minimality for the leading monomial of $f$.  
\end{proof}

Lemma~\ref{lemma:finitedimHGVgen} allows us to establish an upper bound on the maximum degree allowed for any polynomial in $H_{\calS_n}(\R^n \otimes \R^d)$. This is the reason because $\calO(V) = \calO(V)\calO_+(V)^{\calS_n} \oplus H_{\calS_n}(V)$ where $V = \R^n \otimes \R^d$, and $H_{\calS_n}(\R^n \otimes \R^d)$ is a homogenous subspace by Lemma~\ref{lemma:homsub}(c). This immediately implies that $H_{\calS_n}(\R^n \otimes \R^d)$ is finite dimensional as a $\R$-vector space where every subspace $H_{\calS_n}(\R^n \otimes \R^d)^{(j)}$ is spanned by monomials of degree $j$ for $j \leq \binom{n}{2}$. This result is recorded in Lemma~\ref{lem:finite-dim-HSn}. A couple of remarks about Lemma~\ref{lemma:finitedimHGVgen} is needed:
\begin{enumerate}[label=(\roman*)]
    \item The maximum degree $\binom{n}{2}$ of a polynomial in $H_{\calS_n}(\R^n \otimes \R^d)$, does not depend on $d$. 
    
    \item Lemma~\ref{lemma:finitedimHGVgen} also holds for any polynomial $f$ whose lowest degree monomial is greater than $\binom{n}{2}$. This can be shown by expressing $f$ in terms of its homogeneous components.
\end{enumerate}

\begin{lemma}
\label{lem:finite-dim-HSn}
$H_{\calS_n}(\R^n \otimes \R^d)$ is a finite dimensional $\R$-vector space. The maximum degree of a polynomial in $H_{\calS_n}(\R^n \otimes \R^d)$ is at most $\binom{n}{2}$.
\end{lemma}

\subsection{Module structure of \texorpdfstring{$n$}{}-antisymmetric polynomials in terms of \texorpdfstring{$H_{\calS_n}(\R^n \otimes \R^d)$}{}} \label{subsec:modstrantisym}

In this section, we establish a connection between the set of $n$-antisymmetric polynomials $\calO(\R^n \otimes \R^d)_{\chi}$ as a $\calO(\R^n \otimes \R^d)^{\calS_n}$-module and the subspace $H_{\calS_n}(\R^n \otimes \R^d)$. This will allow us to quantify the minimum number of module generators of $\calO(\R^n \otimes \R^d)_{\chi}$. We prove the first three results (Lemma~\ref{lemma:polyvysymhar}, Proposition~\ref{hombasisharmonic}, and Lemma~\ref{lemma:antisymrepsymmod}) in a general setting, where $V$ is an arbitrary vector space, $G$ of $GL(V)$ is a finite group, and $\chi$ is a character of a representation of $G$.
When $V=\R^n \otimes \R^d$, $G = \calS_n$, and $\chi$ the character of the sign representation of $\calS_n$, Lemma~\ref{lemma:antisymrepsymmod} yields the main result of this subsection.

\begin{lemma}
\label{lemma:polyvysymhar}
   For every $k \in \mathbb{N}$, $\calO(V)^{(k)} \subset \calO(V)^G H_G(V)$. Thus, $\calO(V)=\calO(V)^G H_G(V)$.    
\end{lemma}
\begin{proof}
We will use strong induction on $k$ for the proof. If $k=0$, then the assertion is true since the constant polynomials are contained in $H_G(V)$ by definition. Assume the assertion holds for all $k \leq l$. Since $\calO(V) = \calO(V)\calO_+(V)^G \oplus H_G(V)$, by comparing homogeneous parts of degree $l+1$ on both sides, we can say $\calO(V)^{(l+1)}=\big(\calO(V)\calO_+(V)^G\big)^{(l+1)} \oplus H_G(V)^{(l+1)}$. Now, $H_G(V)^{(l+1)} \subset \calO(V)^G H_G(V)$ as $H_G(V)$ is homogenous by Lemma~\ref{lemma:homsub}(c), so we need to show that $\big(\calO(V)\calO_+(V)^G\big)^{(l+1)} \subset \calO(V)^G H_G(V)$. Now clearly $\big(\calO(V)^{(k)}\calO_+(V)^G\big)^{(l+1)}=\{0\}$ if $k>l$ since minimum degree of any polynomial in $\calO_+(V)^G$ is one. Therefore we conclude
\begin{align*}
    \big(\calO(V)\calO_+(V)^G\big)^{(l+1)} =  \sum_{k \leq l} \big(\calO(V)^{(k)}\calO_+(V)^G\big)^{(l+1)} \overset{(a)}{\subset} \sum_{k \leq l} \calO(V)^G\calO(V)^{(k)} \overset{(b)}{\subset} \calO(V)^G H_G(V).
\end{align*}
In the equation above, $\big(\calO(V)^{(k)} \calO_+(V)^G\big)^{(l+1)} = \calO(V)^{(k)} \big(\calO_+(V)^G\big)^{(j)}$ with $j+k=l+1$. Hence, for all $j \leq l+1$, $(\calO_+(V)^G)^{(j)} \subset \calO_+(V)^G \subset \calO(V)^G$ because $\calO_+(V)^G$ and $\calO(V)^G$ are both homogeneous subspaces by Lemma~\ref{lemma:homsub}(c). This explains the inclusion $(a)$. Then $(b)$ follows from the inductive hypothesis. Hence, we are done. 
\end{proof}

\begin{proposition}
\label{hombasisharmonic}
$H_{G}(V) \cap \calO(V)_{\chi}$ as an $\R$-vector space has a homogeneous polynomial basis. 
\end{proposition}

\begin{proof}

By Lemma \ref{lemma:homsub}(c), both $H_G(V)$ and $\calO(V)_{\chi}$ are homogeneous subspaces of $\calO(V)$, which means $H_G(V) \cap \calO(V)_{\chi}$ is also homogeneous. Then $\phi \in H_G(V) \cap \calO(V)_{\chi}$ if and only if every homogeneous component of $\phi$ is in the intersection. Since every vector space has a basis, let $\mathcal{V} = \{\phi_i\}_{i \in \mathcal{I}}$ (for some index set $\mathcal{I}$)  be a basis of $H_G(V) \cap \calO(V)_{\chi}$, and let $\{\psi_{ij}\}_{i \in \mathcal{I}, j \in \N}$ be the set of all homogeneous components of all the elements in $\mathcal{V}$, i.e. $\psi_{ij} = \phi_i^{(j)}$. Then it is clear that $\{\psi_{ij}\}_{i \in \mathcal{I}, j \in \N}$ is a spanning set for $H_G(V) \cap \calO(V)_{\chi}$. Since every spanning set of a vector space contains a basis, our proposition is proven.   
\end{proof}


\vspace*{-0.5cm}
\begin{lemma}
\label{lemma:antisymrepsymmod}
If $\chi$ is the character of a one-dimensional, real representation of $G$, then $\calO(V)_{\chi}=\calO(V)^G(H_G(V) \cap \calO(V)_{\chi})$. In other words, $H_G(V) \cap \calO(V)_{\chi}$ generates $O(V)_{\chi}$ as an $O(V)^G$-module.
\end{lemma}

\begin{proof}
Let $P_{\chi}: \calO(V) \to \calO(V)_{\chi}$ be the linear operator given by
\begin{align*}
    (P_{\chi}f)(v) = \frac{1}{|G|} \sum_{s \in G} \chi(s)f(s.v), \;\; f \in \calO(V), v \in V.
\end{align*}
Our first goal is to prove that $P_{\chi}$ is a projection operator. To make sure $P_{\chi}$ maps $\calO(V)$ into $\calO(V)_{\chi}$, we need $\chi$ to be the character of a one-dimensional, real representation $\rho$ of $G$, because then we have $\chi(g) \coloneqq \tr(\rho(g))=\rho(g)$ for all $g \in G$. Given that $G$ is a finite group, any $g \in G$ must have finite order, and hence $\rho(g)$ must be a real root of unity. So we must have $\rho(g)=\chi(g)=\chi(g^{-1})=\rho(g^{-1})$. In that case, for any $g \in G$, we have for all $v \in V$,
\begin{align*}
    (P_{\chi}f)(g.v) &= \frac{1}{|G|} \sum_{s \in G} \chi(s)f(s.g.v)= \frac{1}{|G|} \sum_{h \in G} \chi(hg^{-1})f(h.v)= \frac{1}{|G|} \sum_{h \in G} \rho(hg^{-1})f(h.v) \\&= \frac{1}{|G|} \sum_{h \in G} \rho(h)\rho(g^{-1})f(h.v)= \frac{\rho(g)}{|G|} \sum_{h \in G} \chi(h)f(h.v) = \chi(g) (P_{\chi}f)(v).
\end{align*}
Next, if $f \in \calO(V)_{\chi}$, then $(P_{\chi}f)(v)=\frac{1}{|G|} \sum_{s \in G} \chi(s)f(s.v)=\frac{1}{|G|} f(v) \sum_{s \in G} \chi(s)^2 = f(v)$. The last equality follows as $\chi(s)$ is a real root of unity, giving us $\chi(s)^2=1$. This completes the proof for $P_{\chi}$ being a projection operator. 

Next, $P_{\chi}$ commutes with any $G$-invariant, constant coefficient differential operator on $\calO(V)$, so $P_{\chi}H_G(V)=H_G(V) \cap \calO(V)_{\chi}$. By Lemma \ref{lemma:polyvysymhar}, any $f \in \calO(V)_{\chi}$ can be written as $f=\sum_{i=1}^M u_ih_i$, where $u_i \in \calO(V)^G$ and $h_i \in H_G(V)$ for all $i$. Applying $P_{\chi}$, we have for all $v \in V$,
\begin{align*}
    f(v)=(P_{\chi}f)(v)&= \frac{1}{|G|} \sum_{s \in G} \chi(s)f(s . v)=\frac{1}{|G|} \sum_{s \in G} \chi(s)\left( \sum_{i=1}^M u_i(s . v) h_i(s . v) \right) \\
    &=\frac{1}{|G|} \sum_{s \in G} \chi(s)\left( \sum_{i=1}^M u_i(v) h_i(s . v) \right)=\sum_{i=1}^M u_i(v) \left(\frac{1}{|G|} \sum_{s \in G} \chi(s)  h_i(s . v) \right)\\
    &= \sum_{i=1}^M u_i(v) (P_{\chi}h_i)(v),
\end{align*}
which completes the proof. 
\end{proof}

It is worth noting that Lemma~\ref{lemma:antisymrepsymmod} is also true when $\chi$ is the character of a one-dimensional, complex representation of $G$ and $V$ is $\C^n \otimes \C^d$. In that case, we define $P_{\chi}$ as $P_{\chi}(f(v)) = \frac{1}{|G|} \sum_{s \in G} \overline{\chi(s)}f(s \cdot v)$. Similarly, we will have $\overline{\chi(g)}=\chi(g^{-1})$, and for $f \in \calO(V)_{\chi}$, we get $P_{\chi}(f(v))=\frac{1}{|G|} f(v) \sum_{s \in G} |\chi(s)|^2 =f(v)$.

Let us specialize the setting to our context: $V=\R^n \otimes \R^d$, $G = \calS_n$, and $\chi=\sgn$. In Lemma~\ref{lem:finitegenatisymonsym}, we prove that $\calO(V)_{\sgn}$ is a finitely generated $\calO(V)^{\calS_n}$-module. Then in Lemma~\ref{lemma:antisymrepsymmod}, we show that $H_{\calS_n}(V) \cap \calO(V)_{\sgn}$ generates $O(V)_{\sgn}$ as an $O(V)^{\calS_n}$-module. Thus, any basis of $H_{\calS_n}(V) \cap \calO(V)_{\sgn}$ (as a $\R$-vector space) can be taken to be the module generators of $\calO(V)_{\sgn}$, and this basis can be taken to be homogenous by Proposition~\ref{hombasisharmonic}. Furthermore, Lemma~\ref{lem:finite-dim-HSn} shows that this basis is finite dimensional, which agrees with Lemma~\ref{lem:finitegenatisymonsym}. Thus, the dimension of the $\R$-vector space $H_{\calS_n}(V) \cap \calO(V)_{\sgn}$ gives us an upper bound on the minimum number of module generators for $\calO(V)_{\sgn}$ as an $O(V)^{\calS_n}$-module. In fact, as the following lemma shows, it is exactly equal to the minimum number of module generators.

\begin{lemma}
\label{lemma:minimalgenasvectspacebasis}
Let $V=\R^n \otimes \R^d$, $G = \calS_n$ and $\chi=\sgn$. Then the minimum number of generators needed to generate $\calO(V)_{\sgn}$ as a module over $\calO(V)^{\calS_n}$ is $\dimension(H_{\calS_n}(V)\cap\calO(V)_{\sgn})$. 
\end{lemma}
\begin{proof}
If $\{f_1, \ldots, f_r\}$ is a minimum sized set of generators for $\calO(V)_{\sgn}$ as an $\calO(V)^{\calS_n}$-module, then by the minimality assumption, they must be linearly independent. Let $W$ be their linear span. 
We will first show that $\calO(V)_{\sgn}=W + \calO(V)^{\calS_n}_+ W$. Suppose $\phi \in \calO(V)_{\sgn}$. Then $\phi=\sum_{i=1}^r u_i f_i$, where $u_i \in \calO(V)^{\calS_n}$ for every $i$. We write each of the $u_i$ in terms of its homogeneous components as $u_i= \sum_{j \geq 0} u_i^{(j)}$. Hence, we get $\phi=\sum_{i=1}^r u_i^{(0)} f_i + \sum_{i=1}^r (\sum_{j \geq 1} u_i^{(j)}) f_i$; in the expression, the first term is in $W$, and the second term is in $\calO(V)^{\calS_n}_+ W$, because $u_i^{(j)} \in \calO(V)^{\calS_n}$ for all $i\in [r]$, and $j \geq 1$ by Lemma~\ref{lemma:homsub}(c). One can easily verify that $\calO(V)^{\calS_n}_+ W = \calO(V)^{\calS_n}_+\calO(V)_{\sgn}$. Thus,
\begin{equation}
\label{eq:upper-bound-quotient}
\dimension(\calO(V)_{\sgn}/\calO(V)^{\calS_n}_+\calO(V)_{\sgn}) \leq r.
\end{equation}
We next claim that 
\begin{equation}
\label{eq:direct-sum-OVsgn}
 \calO(V)_{\sgn} = \calO(V)^{\calS_n}_+\calO(V)_{\sgn} \oplus H_{\calS_n}(V)\cap\calO(V)_{\sgn}.
\end{equation}
We prove this claim at the end, but notice that this claim implies the result. This is true because from \eqref{eq:upper-bound-quotient} and \eqref{eq:direct-sum-OVsgn}, we have $\dimension(H_{\calS_n}(V)\cap\calO(V)_{\sgn}) \leq r$; however, also by the minimality assumption on $\{f_1,\dots,f_r\}$, we get from the comments immediately preceding this lemma, that $r \leq \dimension(H_{\calS_n}(V)\cap\calO(V)_{\sgn})$.

It remains to prove the claim \eqref{eq:direct-sum-OVsgn}. We first note that $\calO(V)^{\calS_n}_+\calO(V)_{\sgn}$ is a subspace of $\calO(V)_{\sgn}$. Let $Z$ be the subspace of $\calO(V)_{\sgn}$ orthogonal to $\calO(V)^{\calS_n}_+\calO(V)_{\sgn}$ with respect to the inner product $\langle \cdot , \cdot \rangle_{\calO(V)}$, so that we have the direct sum decomposition $\calO(V)_{\sgn} = \calO(V)^{\calS_n}_+\calO(V)_{\sgn} \oplus Z$. We will show that $Z = H_{\calS_n}(V)\cap\calO(V)_{\sgn}$. Let us make two important observations:
\begin{enumerate}[label=(\roman*)]
    \item For any $p \in \calO(V)$, let us denote $D_p$ as the constant coefficient differential operator obtained by replacing each $x_{ij}$ in $p$ by $\frac{\partial }{\partial x_{ij}}$ for all $i \in [n]$, $j \in [d]$. Then for all $f,g \in \calO(V)$, we have $\langle D_p f, g \rangle_{\calO(V)} = \langle f, pg \rangle_{\calO(V)}$ for all $p \in \calO(V)$ \citep[cf.][Lemma~3.105]{Wallachbook2017}. 
    \item By Lemma~\ref{lem:Dalpha}, any $p \in \calO(V)^{\calS_n}_+$ and $f \in \calO(V)_{\sgn}$, $D_p f \in \calO(V)_{\sgn}$.
\end{enumerate}
Now, for all $f \in Z$ and $p \in \calO(V)^{\calS_n}_+$, $\langle D_p f, D_p f \rangle_{\calO(V)} = \langle f, p D_p f \rangle_{\calO(V)}= 0$ using observations (i) and (ii); hence, $D_p f = 0$, or equivalently $f \in H_{\calS_n}(V)\cap\calO(V)_{\sgn}$. This shows $Z \subseteq H_{\calS_n}(V)\cap\calO(V)_{\sgn}$. For the converse, assume that $f \in H_{\calS_n}(V)\cap\calO(V)_{\sgn}$, and we want to show $f \in Z$. Take any $g \in \calO(V)^{\calS_n}_+ \calO(V)_{\sgn}$, i.e. it can be expressed as a finite sum $g = \sum_{i} u_i g_i$, where $u_i \in \calO(V)^{\calS_n}_+$ and $g_i \in \calO(V)_{\sgn}$ for every $i$. By observation (i),
\[
    \langle f, g \rangle_{\calO(V)} 
    = \sum_i \langle f, u_i g_i \rangle_{\calO(V)} 
    = \sum_i \langle D_{u_{i}}f, g_i \rangle_{\calO(V)} 
    = 0,
\]
where the last equality follows because $u_i \in \calO(V)^{\calS_n}_+$ implies that $D_{u_i}$ is an $\calS_n$-invariant, constant coefficient differential operator that annihilates polynomials in $H_{\calS_n}(V)$. This proves $f \in Z$.
\end{proof}

Finally, we establish the main theorem using the results in this appendix:

\begin{theorem}
\label{thm:rep-antisymm-polys}
Let $V=\R^n \otimes \R^d$, $G = \calS_n$, and $\chi$ be the sign representation of $\calS_n$. Then $\calO(V)_{\sgn}=\calO(V)^{\calS_n}(H_{\calS_n}(V) \cap \calO(V)_{\sgn})$, i.e. $H_{\calS_n}(V) \cap \calO(V)_{\sgn}$ generates $O(V)_{\sgn}$ as an $O(V)^{\calS_n}$-module. The maximum degree of any polynomial in $H_{\calS_n}(V) \cap \calO(V)_{\sgn}$ is $\binom{n}{2}$. The minimum number of module generators of $\calO(V)_{\sgn}$ is equal to $\dim(H_{\calS_n}(V) \cap \calO(V)_{\sgn})$, as a $\R$-vector space, and the module generators can be chosen to be a homogenous polynomial basis of $H_{\calS_n}(V) \cap \calO(V)_{\sgn}$.
\end{theorem}

\bibliography{antisym1.bib}

\end{document}